\documentclass[twoside,journal]{IEEEtran}
\usepackage{amsmath,amsfonts}
\usepackage{algorithmic}
\usepackage{algorithm}
\usepackage{array}
\usepackage{textcomp}
\usepackage{stfloats}
\usepackage{url}
\usepackage{verbatim}
\usepackage{graphicx}
\usepackage{cite}
\usepackage{booktabs}
\usepackage{xcolor}
\usepackage{subfigure}
\usepackage{rotating}
\usepackage{multirow}
\usepackage{url}
\usepackage[hidelinks]{hyperref}
\usepackage{placeins}
\begin{document}

\title{Spatial-Frequency Enhanced Mamba for Multi-Modal Image Fusion}
\vspace{-4mm}
\author{Hui~Sun,
        Long Lv,
        Pingping~Zhang,
        Tongdan Tang,
        Feng~Tian,
        Weibing Sun,
        and~Huchuan~Lu

\thanks{
Hui~Sun and Long Lv have equal contributions and are co-first authors.
(Corresponding authors: Pingping Zhang and Tongdan Tang.)

Hui~Sun and Pingping~Zhang are with the School of Future Technology, School of Artificial Intelligence, Dalian University of Technology. (Email: sunhui1216@mail.dlut.edu.cn; zhpp@dlut.edu.cn)

Long Lv, Feng Tian and Weibing Sun are with the Affiliated Zhongshan Hospital of Dalian University. (Email: lvlong113@126.com; tianfeng73@163.com; massurm@163.com)

Tongdan Tang is with the Central Hospital of Dalian University of Technology. (Email: tangtongdan2002@sina.com)

Huchuan~Lu is with the School of Information and Communication Engineering, Dalian University of Technology. (Email: lhchuan@dlut.edu.cn)
}
}
\markboth{IEEE Transactions on Image Processing}
{Sun \MakeLowercase{\textit{et al.}}: Spatial-Frequency Enhanced Mamba for Multi-Modal Image Fusion}
\maketitle
\begin{abstract}
Multi-Modal Image Fusion (MMIF) aims to integrate complementary image information from different modalities to produce informative images.
Previous deep learning-based MMIF methods generally adopt Convolutional Neural Networks (CNNs) or Transformers for feature extraction.
However, these methods deliver unsatisfactory performances due to the limited receptive field of CNNs and the high computational cost of Transformers.
Recently, Mamba has demonstrated a powerful potential for modeling long-range dependencies with linear complexity, providing a promising solution to MMIF.
Unfortunately, Mamba lacks full spatial and frequency perceptions, which are very important for MMIF.
Moreover, employing Image Reconstruction (IR) as an auxiliary task has been proven beneficial for MMIF.
However, a primary challenge is how to leverage IR efficiently and effectively.
To address the above issues, we propose a novel framework named Spatial-Frequency Enhanced Mamba Fusion (SFMFusion) for MMIF.
More specifically, we first propose a three-branch structure to couple MMIF and IR, which can retain complete contents from source images.
Then, we propose the Spatial-Frequency Enhanced Mamba Block (SFMB), which can enhance Mamba in both spatial and frequency domains for comprehensive feature extraction.
Finally, we propose the Dynamic Fusion Mamba Block (DFMB), which can be deployed across different branches for dynamic feature fusion.
Extensive experiments show that our method achieves better results than most state-of-the-art methods on six MMIF datasets.
The source code is available at \url{https://github.com/SunHui1216/SFMFusion}.
\end{abstract}
\section{Introduction}
\IEEEPARstart{M}{ulti-Modal} Image Fusion (MMIF) aims to integrate complementary information from different modalities captured by various sensors in the same scene.
Typical MMIF tasks encompass Infrared and Visible Image Fusion (IVIF) as well as Medical Image Fusion (MIF).
In the context of IVIF, infrared images can capture thermal radiation from targets under extreme conditions, such as adverse weather or occlusion, but lack detailed textures.
Conversely, visible images can provide rich texture information but are highly sensitive to changes in illumination and occlusion~\cite{zhang2023visible}.
By integrating their complementary strengths, IVIF can produce information-rich fused images and facilitate subsequent high-level visual tasks, such as depth prediction~\cite{li2020ivfusenet}, semantic segmentation~\cite{jie2024tsjnet,wan2025sigma}, object detection~\cite{liu2022target,zhou2025madinet} and video tracking~\cite{chen2025hyperspectral}.
In the realm of MIF, medical images from diverse modalities capture unique characteristics of human tissues and organs.
For example, Magnetic Resonance Imaging (MRI) provides high-resolution anatomical details of soft tissues, while Computed Tomography (CT) emphasizes dense structural information such as bones and implants~\cite{liu2019medical}.
By leveraging their complementary information, MIF enables a comprehensive representation for medical applications~\cite{james2014medical}.
\begin{figure}[t]
\centering
\includegraphics[width=\linewidth]{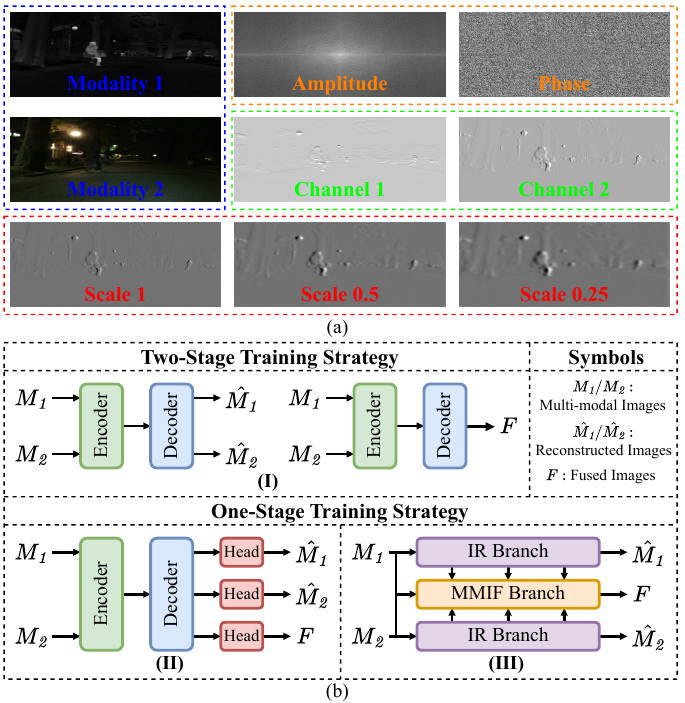}
\vspace{-8mm}
\caption{(a) The multi-modal images and visualized features in the spatial-frequency domains. It can be seen that these features are different and complementary. (b) A comparison between existing IR-assisted MMIF methods (I, II) and our proposed SFMFusion (III).}
\label{fig:introduction}
\vspace{-6mm}
\end{figure}

In recent years, deep learning have significantly enhanced the performance of MMIF.
There are two kinds of outstanding model architectures for extracting image features: Convolutional Neural Networks (CNNs) and Transformers~\cite{vaswani2017attention}.
Compared with traditional handcrafted methods, CNN-based methods~\cite{liu2018infrared,zhong2023fusion,li2021rfn,tang2022image} have demonstrated superior performance.
However, CNNs have a limited receptive field as they extract features primarily from local regions, making it challenging to produce high-quality fused images.
Transformer-based methods~\cite{rao2023tgfuse,vs2022image,zhao2023cddfuse,ma2022swinfusion,tang2023datfuse} have shown promising performance in MMIF.
Nonetheless, the attention mechanism requires quadratic complexity, making it inefficient for MMIF.

Recently, State Space Models (SSMs)~\cite{smith2022simplified,gu2021efficiently,gu2023mamba} have emerged as a promising model architecture in deep learning, and Mamba\cite{gu2023mamba} is a type of SSMs.
Compared with CNNs and Transformers, Mamba achieves a global receptive field with linear complexity.
In fact, several Mamba-based MMIF methods~\cite{li2024mambadfuse,xie2024fusionmamba,ma2024s4fusion,cao2024shuffle,zhu2024mamba} have been developed.
However, these methods overlook a critical issue: Mamba lacks full spatial and frequency perceptions, both of which are essential for MMIF.
Firstly, in the frequency domain, as shown in the first row of Fig.~\ref{fig:introduction}(a), the amplitude spectrum captures the energy distribution across different frequencies, emphasizing the intensity of high-frequency and low-frequency components.
In contrast, the phase spectrum encodes the structural information of the image, highlighting the spatial arrangement of visual features.
Secondly, as shown in the second row of Fig.~\ref{fig:introduction}(a), feature maps in different channels highlight specific characteristics of the image, such as object boundaries.
Thirdly, as shown in the third row of Fig.~\ref{fig:introduction}(a), feature maps at different scales focus on various aspects of the image, ranging from fine-grained details to high-level semantics.
For layout purposes, we rescale the feature maps to the same size.
All these feature maps are distinct and complementary, and they are all crucial for MMIF.

Furthermore, many researches~\cite{li2018densefuse,li2020nestfuse,li2021rfn,erdogan2024fuseformer,zhao2023cddfuse,zhao2024equivariant,tang2023rethinking} show that utilizing Image Reconstruction (IR) as an auxiliary task significantly enhances MMIF.
Current IR-assisted MMIF methods mainly have two paradigms.
The first paradigm is based on the two-stage training strategy (as shown in Fig.~\ref{fig:introduction}(b), (I)).
In this paradigm, IR is performed in the first stage, and the encoder and decoder trained in the first stage are utilized for MMIF in the second stage.
While this two-stage strategy makes full use of IR, it is inefficient.
This is because it is not end-to-end, requiring separate training processes for the two stages.
The second paradigm is based on the one-stage training strategy (as shown in Fig.~\ref{fig:introduction}(b), (II)).
In this paradigm, IR and MMIF are simultaneously performed using multiple task heads.
Although this one-stage strategy is efficient, it does not fully leverage IR.
This is because the features extracted through IR and MMIF are not explicitly disentangled.

To address the aforementioned issues, in this paper, we propose a novel framework named Spatial-Frequency Enhanced Mamba Fusion (SFMFusion) for MMIF.
First of all, we propose a three-branch structure.
It not only performs MMIF but also leverages IR as an auxiliary task, ensuring the retention of complete contents from source images.
Furthermore, we propose the Spatial-Frequency Enhanced Mamba Block (SFMB) to enhance Mamba in both spatial and frequency domains.
As the core part of the entire framework, SFMB can be stacked into groups for comprehensive feature extraction.
Finally, we propose the Dynamic Fusion Mamba Block (DFMB) to achieve a dynamic integration of features from different branches.
Extensive experiments on six public MMIF datasets demonstrate that our framework can attain better performances than most state-of-the-art methods.

In summary, the main contributions are as follows:
\begin{itemize}
 \item{We propose a novel framework named SFMFusion for MMIF, which enhances content preservation through IR.}
 \item{We propose the Spatial-Frequency Enhanced Mamba Block (SFMB) to enhance Mamba in both spatial and frequency domains for comprehensive feature extraction.}
 \item{We propose the Dynamic Fusion Mamba Block (DFMB) to dynamically fuse the features from different branches.}
 \item{Extensive experiments on six public benchmarks demonstrate that our method achieves better performances than most state-of-the-art methods.}
\end{itemize}
\section{Related Work}
\subsection{Multi-Modal Image Fusion}
\textbf{CNN-based Methods.}
In recent years, MMIF has achieved impressive performances with deep learning.
Prior to Transformers, CNN-based methods~\cite{liu2018infrared,zhong2023fusion,li2021rfn,tang2022image} have dominated MMIF.
For example, Liu~\emph{et al.}~\cite{liu2018infrared} use a Siamese CNN to generate a weight map that integrates pixel information from two source images.
Li~\emph{et al.}~\cite{li2018densefuse} introduce dense blocks and employ two fusion strategies for MMIF.
Li~\emph{et al.}~\cite{li2021rfn} propose a feature-based loss function and a residual fusion strategy.
Tang~\emph{et al.}~\cite{tang2022image} cascade an image fusion model with a semantic segmentation model to bridge the gap between MMIF and high-level vision tasks.
However, due to the local receptive field, CNN-based methods struggle to model long-range dependencies, resulting in suboptimal fusion results.

\textbf{Transformer-based Methods.}
Recently, Transformers have been introduced to MMIF.
For example, Vs~\emph{et al.}~\cite{vs2022image} propose a Transformer-based multi-scale fusion strategy.
Rao~\emph{et al.}~\cite{rao2023tgfuse} combine Transformer and adversarial learning for MMIF.
However, the attention mechanism has a quadratic computational complexity.
To reduce the computational complexity, some methods~\cite{ma2022swinfusion,tang2023datfuse} employ Swin Transformer~\cite{liu2021swin} to extract features by leveraging its shifted window mechanism.
Some methods~\cite{zhao2023cddfuse,zhao2024equivariant} use Restormer~\cite{zamir2022restormer} to extract features by computing attention across the channel dimension rather than the spatial dimension.
However, these approaches sacrifice the advantage of the global receptive field.
More recently, Tang~\emph{et~al.}~\cite{tang2024itfuse} propose an interactive Transformer for IVIF.
Yi~\emph{et~al.}~\cite{yi2024text} leverage semantic text guidance for degradation-aware and interactive image fusion.
Wu~\emph{et~al.}~\cite{wu2025fully} propose a fully-connected Transformer for multi-source image fusion.
Wu~\emph{et~al.}~\cite{wang2025wavefusion} propose a wavelet vision Transformer with saliency-guided enhancement for MMIF.
Kang~\emph{et~al.}~\cite{kang2026grformer} build Transformers on Grassmann manifold for IVIF.
These methods highlight promising directions to take advantages of Transformers for image fusion.

\textbf{Mamba-based Methods.}
The newly proposed Mamba~\cite{gu2023mamba} has garnered significant attention due to its remarkable ability to achieve a global receptive field with linear complexity.
Inspired by the success of Mamba, several works~\cite{li2024mambadfuse,xie2024fusionmamba,ma2024s4fusion,cao2024shuffle,zhu2024mamba} have explored its effectiveness in MMIF.
For example, Li~\emph{et al.}~\cite{li2024mambadfuse} leverage Mamba to design a dual-level feature extractor and a dual-phase feature fusion module.
Xie~\emph{et al.}~\cite{xie2024fusionmamba} combine EfficientVMamba~\cite{pei2024efficientvmamba} with two dynamic feature enhancement modules and a cross-modality fusion module.
Zhu~\emph{et al.}~\cite{zhu2024mamba} build a Transformer-Mamba hybrid framework.
However, all the above methods overlook a critical issue: Mamba lacks full spatial and frequency perceptions, both of which are essential for MMIF.
Different from them, which use Mamba as a plug-and-play module, we enhance Mamba in the spatial-frequency domains to extract more comprehensive features.
\begin{figure*}[t]
  \centering
  \includegraphics[width=0.9\textwidth, height=10cm]{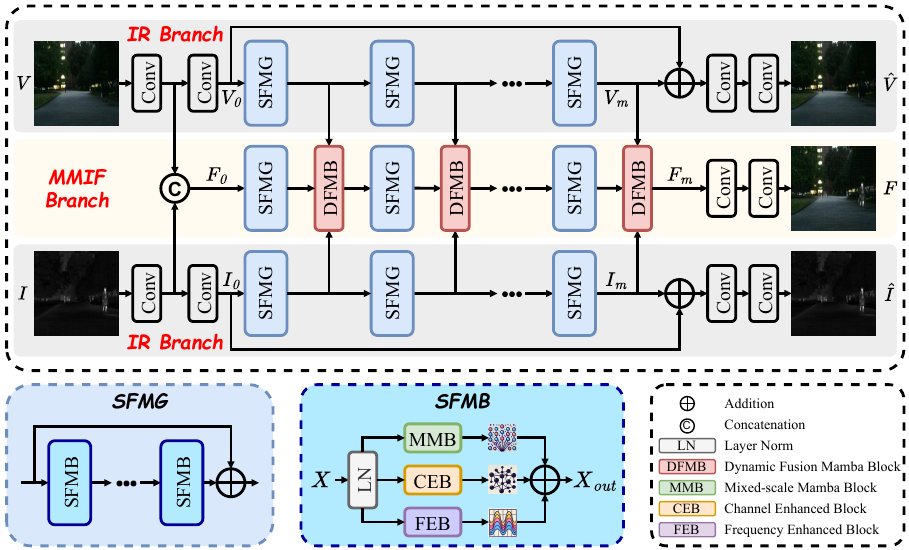}
  \vspace{-2mm}
  \caption{The overall architecture of our proposed SFMFusion. It includes the Spatial-Frequency Enhanced Mamba Group (SFMG) and the Dynamic Fusion Mamba Block (DFMB). A SFMG includes stacked SFMBs and a skip connection. A SFMB includes a Mixed-scale Mamba Block (MMB), a Channel Enhanced Block (CEB) and a Frequency Enhanced Block (FEB).}
  \label{fig:framework}
  \vspace{-4mm}
\end{figure*}
\subsection{State Space Model}
In continuous linear time-invariant systems, State Space Models (SSMs)~\cite{smith2022simplified,gu2021efficiently,gu2023mamba} establish a mapping from \( x(t) \in \mathbb{R} \) to response \( y(t) \in \mathbb{R} \) by the hidden state \( h(t) \in \mathbb{R}^N \).
The system can be mathematically expressed using a linear Ordinary Differential Equation (ODE):
\begin{align}
h'(t) &= Ah(t) + Bx(t), \label{eq:ode1} \\
y(t) &= Ch(t), \label{eq:ode2}
\end{align}
where \( A \in \mathbb{R}^{N \times N} \) is the state matrix, \( B \in \mathbb{R}^{N \times 1} \) is the input matrix and \( C \in \mathbb{R}^{1 \times N} \) is the output matrix.

To incorporate SSMs into deep learning frameworks, the continuous differential equations are discretized.
We can discretize the system in Eq.\ref{eq:ode1} and Eq.\ref{eq:ode2} using the Zero-Order Hold (ZOH), which can be formally defined as follows:
\begin{align}
h_t &= \bar{A} h_{t-1} + \bar{B} x_t, \label{eq:zoh1} \\
y_t &= Ch_t, \label{eq:zoh2}
\end{align}
where \( \bar{A} = \exp(\Delta A) \) and \( \bar{B} = (\Delta A)^{-1} (\exp(\Delta A) - I) \cdot \Delta B \) are the discretized state parameters. \( \Delta \) represents the step size in the discretization progress.

However, the expression presented in Eq.\ref{eq:zoh1} and Eq.\ref{eq:zoh2} concentrates on the linear time-invariant systems with parameters that do not vary with different inputs.
To overcome this constraint, Mamba~\cite{gu2023mamba} integrates a selective scanning mechanism, parameterizing \( B \), \( C \) and \( \Delta \) as functions of the input.
In addition, by relying on a faster hardware-aware algorithm, Mamba has further advanced its potential.
For image processing, VMamba~\cite{liu2024vmamba} introduces the 2D Selective Scan Module (2D-SSM), which flattens 2D images into four 1D sequences and scans along four distinct directions.
\section{Proposed Method}
In this section, we introduce our proposed SFMFusion.
It comprises two key components: Spatial-Frequency Enhanced Mamba Group (SFMG) and Dynamic Fusion Mamba Block (DFMB).
Note that SFMFusion is a general MMIF framework, and we only take IVIF as an example for illustration.
\subsection{Overview of SFMFusion}
As illustrated in Fig.~\ref{fig:framework}, our proposed SFMFusion consists of three branches: two IR branches and one MMIF branch.

In the IR branches, taking the visible image as an example, the input \(V\) is first fed into two 3$\times$3 convolutional layers to generate the initial feature \(V_{\mathit{0}}\).
Subsequently, it undergoes multiple SFMGs.
This process can be expressed as:
\begin{align}
  V_{\mathit{m}} &= \text{SFMG}_{\mathit{m}}(\text{SFMG}_{\mathit{m-1}}(\cdots\text{SFMG}_{\mathit{1}}(V_{\mathit{0}}))),
\end{align}
where \(\mathit{m}\) is the number of SFMGs and \(V_{\mathit{m}}\) is the output of \(\mathit{m}\)-th SFMG.
Afterwards, we incorporate a skip connection to extract high-quality features.
Finally, two 3$\times$3 convolutional layers are employed to obtain the reconstructed image \(\hat{V}\).

In the MMIF branch, the inputs \(V\) and \(I\) are first fed into 3$\times$3 convolutional layers and then concatenated to generate the initial feature \(F_{\mathit{0}}\).
Subsequently, it undergoes multiple SFMGs and DFMBs.
This process can be expressed as:
\begin{align}
  F_{\mathit{m}} &= \text{DFMB}_{\mathit{m}}(\text{SFMG}_{\mathit{m}}(\cdots\text{DFMB}_{\mathit{1}}(\text{SFMG}_{\mathit{1}}(F_{\mathit{0}})))),
\end{align}
where \(F_{\mathit{m}}\) is the output of \(\mathit{m}\)-th DFMB.
Finally, two 3$\times$3 convolutional layers are used to obtain the fused image $F$.
As observed, all of the three branches have some SFMGs.
To improve the representation ability, these SFMGs are trained independently rather than sharing weights.
\subsection{Spatial-Frequency Enhanced Mamba Block}
Most existing Mamba-based methods~\cite{li2024mambadfuse,xie2024fusionmamba,ma2024s4fusion,cao2024shuffle,zhu2024mamba} simply stack Mamba blocks to extract image features, ignoring that Mamba lacks full spatial and frequency perceptions.
To address this issue, we propose the Spatial-Frequency Enhanced Mamba Block (SFMB).
A SFMB includes a Mixed-scale Mamba Block (MMB) for spatial perception in the spatial dimension, a Channel Enhanced Block (CEB) for spatial perception in the channel dimension and a Frequency Enhanced Block (FEB) for frequency perception.
As shown in Fig.~\ref{fig:framework}, given the input \( X \), the output of the SFMB can be defined as:
\begin{align}
\bar{X} &= \text{LN}(X), \label{sfmb1} \\
X_{\mathit{out}} &= \text{MMB}(\bar{X}) + \text{CEB}(\bar{X}) + \text{FEB}(\bar{X}),
\label{sfmb2}
\end{align}
where LN represents the layer normalization~\cite{xiong2020layer}.
\subsubsection{Mixed-scale Mamba Block}
\begin{figure}[!t]
  \centering
  \includegraphics[width=\linewidth]{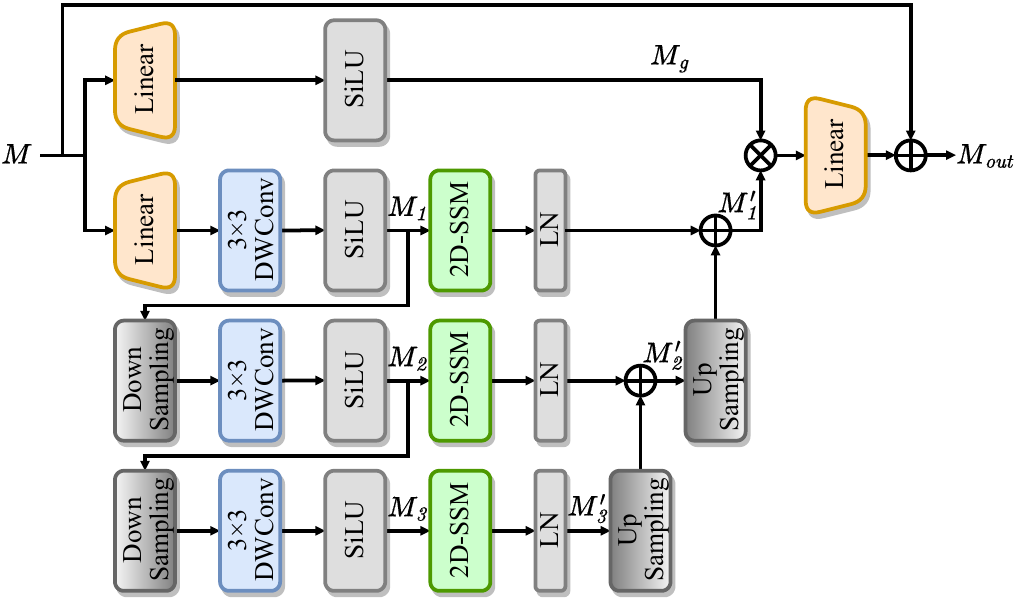}
  \vspace{-4mm}
  \caption{The structure of our proposed MMB.}
  \label{fig:MMB}
  \vspace{-4mm}
\end{figure}
In the MMIF task, both fine-grained details and high-level semantics are crucial.
However, the original Mamba can only extract features at a single scale, which leads to the loss of multi-scale information.
To handle this drawback, we propose MMB to extract features at different scales to enhance Mamba in the spatial dimension.

As shown in Fig.~\ref{fig:MMB}, in the first stream, the input \( M \) is expanded along the channel dimension by a linear layer with a 3$\times$3 depth-wise convolutional layer and a SiLU activation function to generate \(M_{\mathit{1}}\).
Then, a series of operations including a down-sampling layer, a 3$\times$3 depth-wise convolutional layer and a SiLU activation function are sequentially stacked to generate \(M_{\mathit{2}}\). \(M_{\mathit{3}}\) is obtained in the same way as \(M_{\mathit{2}}\).
The process of extracting multi-scale features can be defined as:
\begin{align}
  M_{\mathit{1}} &= \theta(\text{DWConv}_{3 \times 3}(\text{Linear}(M))), \\
  M_{\mathit{2}} &= \theta(\text{DWConv}_{3 \times 3}(\text{Down}(M_{\mathit{1}}))), \\
  M_{\mathit{3}} &= \theta(\text{DWConv}_{3 \times 3}(\text{Down}(M_{\mathit{2}}))),
\end{align}
where Linear, \(\text{DWConv}_{3\times3}\), Down and \( \theta \) are the linear layer, the 3$\times$3 depth-wise convolutional layer, the down-sampling layer and the SiLU activation function~\cite{elfwing2018sigmoid}, respectively.

Afterwards, the 2D-SSM with the layer normalization is utilized as the key module to refine the above features. The up-sampling layers are utilized to restore the features to their original sizes.
The process of refining and restoring multi-scale features can be defined as:
\begin{align}
  M_{\mathit{3}}^{\prime} &= \text{LN}(\text{2D-SSM}(M_3)), \\
  M_{\mathit{2}}^{\prime} &= \text{LN}(\text{2D-SSM}(M_2)) + \text{Up}(M_{\mathit{3}}^{\prime}), \\
  M_{\mathit{1}}^{\prime} &= \text{LN}(\text{2D-SSM}(M_1)) + \text{Up}(M_{\mathit{2}}^{\prime}),
\end{align}
where Up denotes the up-sampling layer.

In the second stream, another linear layer and a SiLU activation function are used.
The output of this stream can be obtained by:
\begin{align}
  M_{\mathit{g}} &= \theta(\text{Linear}(M)).
\end{align}

To obtain the final output \(M_{\mathit{out}}\), \(M_{\mathit{1}}^{\prime}\) and \(M_{\mathit{g}}\) are fused via the element-wise multiplication, followed by a linear layer and a skip connection.
It can be expressed by:
\begin{align}
  M_{\mathit{out}} &= \text{Linear}(M_{\mathit{1}}^{\prime} \cdot M_{\mathit{g}}) + M.
\end{align}

By extracting features at mixed scales, MMB can better capture fine-grained details and high-level semantics, enhancing Mamba in the spatial dimension of the spatial domain.
\subsubsection{Channel Enhanced Block}
\begin{figure}[!t]
  \centering
  \includegraphics[width=\linewidth]{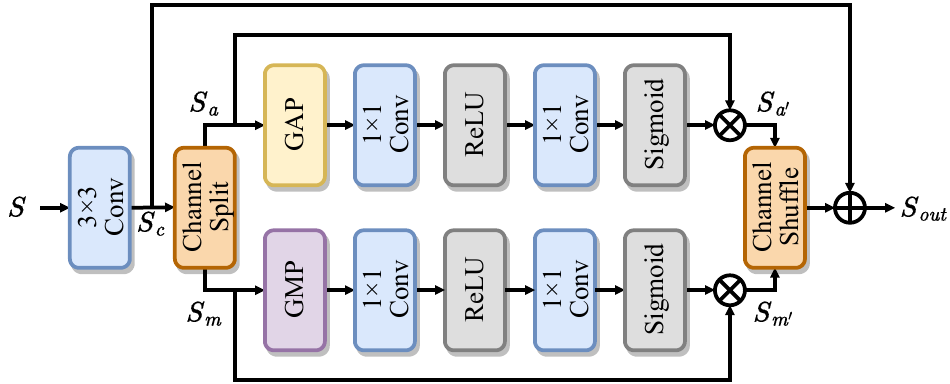}
  \vspace{-4mm}
  \caption{The structure of our proposed CEB.}
  \label{fig:CEB}
  \vspace{-4mm}
\end{figure}
As a low-level visual task, MMIF requires robust modeling of both the spatial and channel dimensions.
However, the scanning of Mamba operates along the height and width directions of the feature, it struggles to model inter-channel information.
To handle this drawback, we propose CEB to enhance Mamba in the channel dimension.

As shown in Fig.~\ref{fig:CEB}, the input feature \(S\) is first processed through a 3$\times$3 convolutional layer to obtain \(S_{\mathit{c}}\).
Then, we split \(S_{\mathit{c}}\) into two parts, \(S_{\mathit{a}}\) and \(S_{\mathit{m}}\), by halving the channel dimension.
\(S_{\mathit{a}}\) and \(S_{\mathit{m}}\) are subsequently sent into the first and second streams, respectively.
Considering that different global pooling methods emphasize distinct regions, the first stream uses Global Average Pooling (GAP) and the second stream uses Global Max Pooling (GMP).
After pooling, both streams adopt the same processing steps: an 1$\times$1 convolutional layer with the ReLU activation function and another 1$\times$1 convolutional layer with the Sigmoid activation function.
The processed features are combined with \(S_{\mathit{a}}\) and \(S_{\mathit{m}}\) through the element-wise multiplication.
\begin{align}
  S_{\mathit{a}^{\prime}} &= S_{\mathit{a}} \cdot \sigma(\text{Conv}_{1 \times 1}(\phi(\text{Conv}_{1 \times 1}(\text{GAP}(S_{\mathit{a}})))))  , \\
  S_{\mathit{m}^{\prime}} &= S_{\mathit{m}} \cdot \sigma(\text{Conv}_{1 \times 1}(\phi(\text{Conv}_{1 \times 1}(\text{GMP}(S_{\mathit{m}})))))  ,
\end{align}
where \(\phi\) and \(\sigma\) denote the ReLU and Sigmoid activation functions, respectively.
Finally, the two enhanced streams are integrated using channel shuffle, combined with a skip connection to generate the output \(S_{\mathit{out}}\).

By considering both the average and maximum responses across channels, CEB can better model channel-wise information, thereby enhancing Mamba in the channel dimension of the spatial domain.
\subsubsection{Frequency Enhanced Block}
It is recognized that the spatial domain predominantly resides in low-frequency information.
In MMIF, low-frequency information such as global contents and high-frequency information such as local details are both important.
However, Mamba lacks the capability to effectively process information in the frequency domain.
To handle this drawback, we propose FEB to enhance Mamba in the frequency domain.
\begin{figure}[!t]
  \centering
  \includegraphics[width=\linewidth]{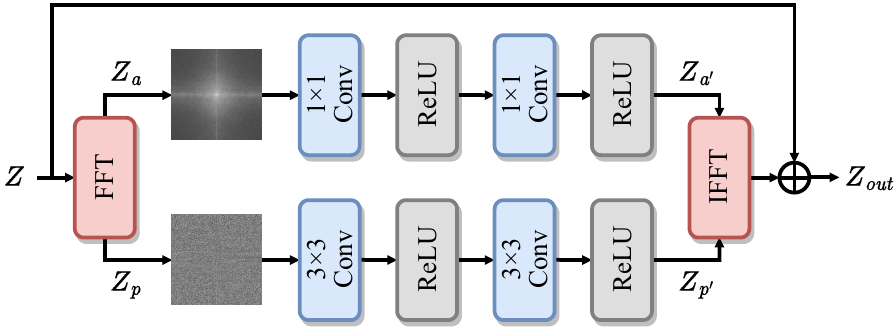}
  \vspace{-4mm}
  \caption{The structure of our proposed FEB.}
  \label{fig:FEB}
  \vspace{-4mm}
\end{figure}

As shown in Fig.~\ref{fig:FEB}, the input \(Z\) is first transformed into its Fourier spectrum via the Fast Fourier Transform (FFT)~\cite{Cooley1965AnAF}, which is decomposed into the amplitude spectrum \(Z_{\mathit{a}}\) and phase spectrum \(Z_{\mathit{p}}\).
The amplitude spectrum contains high-frequency details, while the phase spectrum contains structural information.
Therefore, in the first stream, two 1$\times$1 convolutional layers with the ReLU activation functions are used to yield \(Z_{\mathit{a}^{\prime}}\).
In the second stream, 3$\times$3 convolutional layers are used to yield \(Z_{\mathit{p}^{\prime}}\).
They can be expressed as:
\begin{align}
  Z_{\mathit{a}^{\prime}} &= \phi(\text{Conv}_{1 \times 1}(\phi(\text{Conv}_{1 \times 1}(Z_{\mathit{a}})))), \\
  Z_{\mathit{p}^{\prime}} &= \phi(\text{Conv}_{3 \times 3}(\phi(\text{Conv}_{3 \times 3}(Z_{\mathit{p}})))).
\end{align}

Finally, \(Z_{\mathit{a}^{\prime}}\) and \(Z_{\mathit{p}^{\prime}}\) are transformed to
the spatial domain through the Inverse Fast Fourier Transform (IFFT), combined with a skip connection to generate the output \(Z_{\mathit{out}}\).

By separately processing the amplitude spectra and phase spectra, FEB can better handle different types of information, thereby enhancing Mamba in the frequency domain.
\subsection{Dynamic Fusion Mamba Block}
Since our model is built upon a three-branch structure, it is essential to effectively utilize the features from IR branches to ensure the preservation of complete contents.
Although the addition or concatenation can be utilized, these simple fusion strategies fail to fully utilize the features from IR branches.
This is because such strategies are fixed and do not adapt to the varying characteristics of different modalities.
To address this issue, we introduce DFMB for dynamic feature fusion.
The core idea behind DFMB is to dynamically and adaptively allocate the importance of features coming from the two IR branches, allowing the model to focus on the most essential features during the fusion process.

\begin{figure}[!t]
  \centering
  \includegraphics[width=\linewidth]{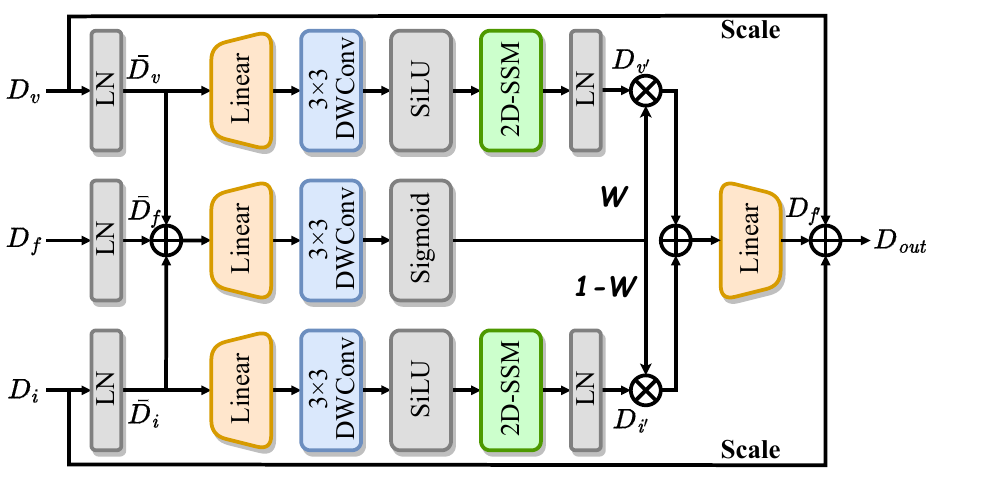}
  \vspace{-4mm}
  \caption{The structure of our proposed DFMB.}
  \label{fig:DFMB}
  \vspace{-4mm}
\end{figure}

As shown in Fig.~\ref{fig:DFMB}, the inputs \(D_{\mathit{v}}\), \(D_{\mathit{f}}\) and \(D_{\mathit{i}}\) are first processed through the layer normalization to obtain \(\bar{D_{\mathit{v}}}\), \(\bar{D_{\mathit{f}}}\) and \(\bar{D_{\mathit{i}}}\).
Then, \(\bar{D_{\mathit{v}}}\) and \(\bar{D_{\mathit{i}}}\) are processed through a series of operations including a linear layer, a 3$\times$3 depth-wise convolutional layer, a SiLU activation function and a 2D-SSM with a layer normalization to obtain \(D_{\mathit{v}^{\prime}}\) and \(D_{\mathit{i}^{\prime}}\).
The weight \( W \) is computed by considering the features from all the three streams by an element-wise addition, a linear layer, a 3$\times$3 depth-wise convolutional layer and a Sigmoid activation function.
The above procedure can be expressed as:
\begin{align}
  D_{\mathit{v}^{\prime}} &= \text{LN}(\text{2D-SSM}(\theta(\text{DWConv}_{3 \times 3}(\text{Linear}(\bar{D_v}))))),\\
  D_{\mathit{i}^{\prime}} &= \text{LN}(\text{2D-SSM}(\theta(\text{DWConv}_{3 \times 3}(\text{Linear}(\bar{D_i}))))),\\
  W &= \sigma(\text{DWConv}_{3 \times 3}(\text{Linear}(\bar{D_i} + \bar{D_f} + \bar{D_v}))).
\end{align}

Finally, \(\bar{D_{\mathit{v}}}\), \(\bar{D_{\mathit{i}}}\) and \( W \) are integrated with a linear layer and two scaled skip connections:
\begin{align}
  D_{\mathit{f}^{\prime}} &= \text{Linear}(D_{\mathit{v}^{\prime}} \cdot W + D_{\mathit{i}^{\prime}} \cdot (1-W)),\\
  D_{\mathit{out}} &= D_{\mathit{f}^{\prime}} + s_1 \cdot D_v + s_2 \cdot D_i,
\end{align}
where \( s_1 \) and \( s_2 \) represent the learnable scale factors.

By dynamically fusing features from IR branches, DFMB can better preserve the complete contents from source images.
\begin{table*}[htpb]
\caption{Quantitative results on the MSRS dataset. The best result is in \textbf{bold} and the second-best is \underline{underlined}.}
\vspace{-2mm}
\label{tab:comparison1}
\centering
\renewcommand{\arraystretch}{1.0} 
\setlength{\tabcolsep}{8pt} 
\begin{tabular}{llcccccccccc}
\toprule
\textbf{Method} & \textbf{Venue} & \textbf{EN} ($\uparrow$) & \textbf{SD} ($\uparrow$) & \textbf{SF} ($\uparrow$) & \textbf{AG} ($\uparrow$) & \textbf{MI} ($\uparrow$) & \textbf{VIF} ($\uparrow$) & \textbf{Q\textsuperscript{AB/F}} ($\uparrow$) & \textbf{Avg.Rank} ($\downarrow$) \\
\midrule
RFN-Nest\cite{li2021rfn}         & IF 21      & 6.18 & 29.04 & 6.20 & 2.14 & 1.67 & 0.64 & 0.38 & 12.86 \\
SDNet\cite{zhang2021sdnet}            & IJCV 21    & 5.26 & 17.32 & 8.69 & 2.70 & 1.14 & 0.49 & 0.37 & 13.00 \\
SeAFusion\cite{tang2022image}        & IF 22      & 6.65 & 41.84 & 11.11 & 3.68 & 2.79 & 0.97 & 0.67 & 4.29 \\
UMF-CMGR\cite{UMF}         & IJCAI 22   & 5.60 & 20.75 & 7.13 & 2.15 & 1.31 & 0.42 & 0.26 & 14.86 \\
SwinFusion\cite{ma2022swinfusion}       & JAS 22     & 6.62 & 43.00 & 11.09 & 3.55 & 3.14 & 0.99 & 0.65 & 4.29 \\
U2Fusion\cite{xu2020u2fusion}         & TPAMI 22   & 5.03 & 20.29 & 7.22 & 2.25 & 1.37 & 0.49 & 0.34 & 14.29 \\
DeFusion\cite{liang2022fusion}         & ECCV 22    & 6.35 & 34.89 & 7.98 & 2.60 & 2.16 & 0.75 & 0.51 & 9.29 \\
ReCoNet\cite{huang2022reconet}          & ECCV 22    & 4.23 & 41.71 & 9.98 & 2.99 & 1.58 & 0.49 & 0.40 & 10.86 \\
IRFS\cite{wang2023interactively}             & IF 23      & 6.17 & 29.79 & 8.34 & 2.66 & 1.82 & 0.74 & 0.53 & 10.00 \\
LRRNet\cite{li2023lrrnet}           & TPAMI 23   & 6.19 & 31.77 & 8.46 & 2.63 & 2.04 & 0.54 & 0.46 & 10.14 \\
DDFM\cite{zhao2023ddfm}             & ICCV 23    & 6.17 & 28.92 & 7.39 & 2.51 & 1.89 & 0.74 & 0.47 & 11.14 \\
SegMif\cite{liu2023multi}           & ICCV 23    & 6.40 & 41.96 & 11.01 & 3.60 & 1.78 & 0.76 & 0.58 & 6.71 \\
CDDFuse\cite{zhao2023cddfuse}          & CVPR 23    & \underline{{6.70}} & 43.37 & \underline{{11.56}} & 3.73 &
\textbf{{3.47}} & \underline{{1.05}} & \underline{{0.69}} & \underline{{2.29}} \\
EMMA\cite{zhao2024equivariant}             & CVPR 24    & \textbf{{6.72}} & \underline{{44.59}} & \underline{{11.56}} & \underline{{3.77}} & 2.94 & 0.97 & 0.64 & 2.86 \\
MMDRFuse\cite{deng2024mmdrfuse}         & ACM MM 24  & 6.47 & 37.32 & 9.64 & 3.22 & 2.49 & 0.84 & 0.58 & 6.71 \\
SFMFusion(Ours)         & -          & \textbf{{6.72}} & \textbf{{45.17}} & \textbf{{11.94}} & \textbf{{3.90}} & \underline{{3.39}} & \textbf{{1.07}} & \textbf{{0.71}} & \textbf{{1.14}} \\
\bottomrule
\end{tabular}
\vspace{-2mm}
\end{table*}
\begin{table*}[htpb]
\caption{Quantitative results on the M3FD dataset. The best result is in \textbf{bold} and the second-best is \underline{underlined}.}
\vspace{-2mm}
\label{tab:comparison2}
\centering
\renewcommand{\arraystretch}{1.0} 
\setlength{\tabcolsep}{8pt} 
\begin{tabular}{llcccccccccc}
\toprule
\textbf{Method} & \textbf{Venue} & \textbf{EN} ($\uparrow$) & \textbf{SD} ($\uparrow$) & \textbf{SF} ($\uparrow$) & \textbf{AG} ($\uparrow$) & \textbf{MI} ($\uparrow$) & \textbf{VIF} ($\uparrow$) & \textbf{Q\textsuperscript{AB/F}} ($\uparrow$) & \textbf{Avg.Rank} ($\downarrow$) \\
\midrule
RFN-Nest\cite{li2021rfn}         & IF 21      & 6.86 & 33.60 & 7.74 & 2.86 & 1.98 & 0.58 & 0.40 & 12.00 \\
SDNet\cite{zhang2021sdnet}            & IJCV 21    & 6.84 & 35.08 & 13.62 & 4.74 & 2.19 & 0.57 & 0.52 & 8.29 \\
SeAFusion\cite{tang2022image}        & IF 22      & 6.85 & 35.43 & 13.95 & 4.77 & 2.48 & 0.72 & 0.60 & 5.43 \\
UMF-CMGR\cite{UMF}         & IJCAI 22   & 6.70 & 30.53 & 8.78 & 2.94 & 2.11 & 0.61 & 0.39 & 11.71 \\
SwinFusion\cite{ma2022swinfusion}       & JAS 22     & 6.79 & 35.84 & 13.69 & 4.60 & \underline{{2.88}} & 0.77 & \underline{{0.61}} & 4.86 \\
U2Fusion\cite{xu2020u2fusion}         & TPAMI 22   & 6.64 & 28.73 & 10.68 & 3.94 & 1.90 & 0.63 & 0.54 & 10.57 \\
DeFusion\cite{liang2022fusion}         & ECCV 22    & 6.46 & 27.16 & 7.38 & 2.58 & 2.05 & 0.54 & 0.33 & 15.00 \\
ReCoNet\cite{huang2022reconet}          & ECCV 22    & 6.63 & 34.36 & 10.55 & 3.92 & 2.11 & 0.59 & 0.49 & 10.43 \\
IRFS\cite{wang2023interactively}             & IF 23      & 6.69 & 30.25 & 10.15 & 3.43 & 1.98 & 0.66 & 0.55 & 10.29 \\
LRRNet\cite{li2023lrrnet}           & TPAMI 23   & 6.44 & 27.17 & 10.68 & 3.59 & 1.96 & 0.57 & 0.50 & 12.43 \\
DDFM\cite{zhao2023ddfm}             & ICCV 23    & 6.72 & 30.48 & 9.12 & 3.17 & 2.26 & 0.61 & 0.46 & 10.71 \\
SegMif\cite{liu2023multi}           & ICCV 23    & \textbf{{6.98}} & 37.68 & 14.25 & 4.81 & 2.11 & \textbf{{0.80}} & \textbf{{0.65}} & 3.29 \\
CDDFuse\cite{zhao2023cddfuse}          & CVPR 23    & 6.90 & 37.24 & 14.78 & 4.86 & 2.71 & \underline{{0.79}} & \underline{{0.61}} & \underline{{3.00}} \\
EMMA\cite{zhao2024equivariant}             & CVPR 24    & 6.92 & \underline{{38.26}} & \textbf{{15.23}} & \textbf{{5.33}} & 2.64 & 0.77 & 0.59 & \underline{{3.00}} \\
MMDRFuse\cite{deng2024mmdrfuse}         & ACM MM 24  & 6.40 & 26.06 & 10.00 & 3.35 & 2.25 & 0.66 & 0.51 & 11.43 \\
SFMFusion(Ours)  & -          & \underline{{6.93}} & \textbf{{38.65}} & \underline{{14.81}} & \underline{{4.88}} & \textbf{{2.92}} & \underline{{0.79}} & \underline{{0.61}} & \textbf{{1.71}} \\
\bottomrule
\end{tabular}
\vspace{-2mm}
\end{table*}
\subsection{Loss Functions}
The total loss \( \mathcal{L}_{\mathit{total}} \) comprises three parts: the fusion loss \( \mathcal{L}_{\mathit{f}} \), the visible image reconstruction loss \( \mathcal{L}_{\mathit{v}} \) and the infrared image reconstruction loss \( \mathcal{L}_{\mathit{i}} \).
More specifically, the total loss \( \mathcal{L}_{\mathit{total}} \) can be expressed as follows:
\begin{equation}
\mathcal{L}_{\mathit{total}} = \mathcal{L}_{\mathit{f}} + \alpha_1 \mathcal{L}_{\mathit{v}} + \alpha_2 \mathcal{L}_{\mathit{i}}, \label{eq:total_loss}
\end{equation}
where \( \alpha_1 \) and \( \alpha_2 \) are the hyper-parameters to balance the loss terms.
\( \mathcal{L}_{\mathit{f}} \) is composed of \( \mathcal{L}_{\mathit{int}} \) and \( \mathcal{L}_{\mathit{grad}} \), as follows:
\begin{equation}
\mathcal{L}_{\mathit{f}} = \mathcal{L}_{\mathit{int}} + \alpha_3 \mathcal{L}_{\mathit{grad}}, \label{eq:fusion_loss}
\end{equation}
where \( \alpha_3 \) is the hyper-parameter to balance the loss terms.

\begin{equation}
\mathcal{L}_{\mathit{int}} = \frac{1}{HW} \left\| F - \max(I, V) \right\|_1 \label{eq:int_loss},
\end{equation}
where \( H \) and \( W \) are the height and width of an image. \( \|\cdot\|_1 \) stands for the \( l_1 \)-norm and \( \max(\cdot) \) denotes the element-wise maximum selection.

\begin{equation}
\mathcal{L}_{\mathit{grad}} = \frac{1}{HW} \left\| \nabla F - \max(|\nabla I|, |\nabla V|) \right\|_1 \label{eq:grad_loss},
\end{equation}
where \( \nabla \) indicates the Sobel gradient operator and \( |\cdot| \) refers to the absolute operation.

The calculations for \( \mathcal{L}_{\mathit{v}} \) and \( \mathcal{L}_{\mathit{i}} \) are the same. Taking \( \mathcal{L}_{\mathit{v}} \) as an example:
\begin{align}
\mathcal{L}_{\mathit{v}} &= \mathcal{L}_{\mathit{int}} + \alpha_4 \mathcal{L}_{\mathit{grad}}, \label{eq:re_loss} \\
\mathcal{L}_{\mathit{int}} &= \frac{1}{HW} \left\| \hat{V} - V \right\|_1,\label{eq:re_int_loss} \\
\mathcal{L}_{\mathit{grad}} &= \frac{1}{HW} \left\| \nabla \hat{V} - |\nabla V| \right\|_1, \label{eq:re_grad_loss}
\end{align}
where \( \alpha_4 \) is the hyper-parameter to balance the loss terms.
\section{Experiments}
\subsection{Datasets}
In this paper, we conduct extensive experiments to evaluate our method on six public MMIF datasets.

For IVIF, we adopt three datasets: MSRS~\cite{tang2022piafusion}, M3FD~\cite{liu2022target} and FMB~\cite{liu2023multi}.
For training, we adopt the MSRS training set, which contains 1,083 image pairs.
For testing, we utilize the MSRS testing set (361 pairs), the M3FD testing set (300 pairs) and the FMB testing set (280 pairs), allowing for a thorough performance assessment.
These datasets include a diverse range of images captured in both the day and night, and include various objects such as people, cars and bikes.

For MIF, we adopt the Harvard Medical dataset\footnote{\url{http://www.med.harvard.edu/AANLIB/home.html}}.
Following previous works~\cite{zhao2023cddfuse,zhao2024equivariant}, we use 21 pairs of MRI-CT images, 42 pairs of MRI-PET images and 73 pairs of MRI-SPECT images for testing.
These selected pairs represent diverse medical modalities and complex structures of organs, allowing for a comprehensive evaluation across diverse MIF tasks.

It is important to note that no fine-tuning is applied to the M3FD, FMB and Harvard Medical dataset to demonstrate the generalization capability of our proposed model.
\begin{table*}[htpb]
\caption{Quantitative results on the FMB dataset. The best result is in \textbf{bold} and the second-best is \underline{underlined}.}
\vspace{-2mm}
\label{tab:comparison3}
\centering
\renewcommand{\arraystretch}{1.0} 
\setlength{\tabcolsep}{8pt} 
\begin{tabular}{llcccccccccc}
\toprule
\textbf{Method} & \textbf{Venue} & \textbf{EN} ($\uparrow$) & \textbf{SD} ($\uparrow$) & \textbf{SF} ($\uparrow$) & \textbf{AG} ($\uparrow$) & \textbf{MI} ($\uparrow$) & \textbf{VIF} ($\uparrow$) & \textbf{Q\textsuperscript{AB/F}} ($\uparrow$) & \textbf{Avg.Rank} ($\downarrow$) \\
\midrule
RFN-Nest\cite{li2021rfn}         & IF 21      & \underline{{6.82}} & 35.32 & 7.81  & 2.57 & 2.14 & 0.61 & 0.46 & 10.86 \\
SDNet\cite{zhang2021sdnet}            & IJCV 21    & 6.61 & 34.80 & 13.11 & 4.04 & 2.13 & 0.58 & 0.55 & 10.14 \\
SeAFusion\cite{tang2022image}        & IF 22      & 6.75 & 36.16 & 13.88 & 4.26 & 2.69 & 0.80 & 0.65 & 5.29 \\
UMF-CMGR\cite{UMF}         & IJCAI 22   & 6.55 & 30.24 & 8.73  & 2.47 & 2.17 & 0.62 & 0.41 & 12.86 \\
SwinFusion\cite{ma2022swinfusion}       & JAS 22     & 6.67 & 35.38 & 13.51 & 4.04 & \textbf{{3.08}} & 0.85 & 0.67 & 4.71 \\
U2Fusion\cite{xu2020u2fusion}         & TPAMI 22   & 6.57 & 29.05 & 10.27 & 3.40 & 2.10 & 0.66 & 0.58 & 10.71 \\
DeFusion\cite{liang2022fusion}         & ECCV 22    & 6.39 & 27.96 & 7.13  & 2.22 & 2.29 & 0.59 & 0.37 & 13.86 \\
ReCoNet\cite{huang2022reconet}          & ECCV 22    & 6.69 & 33.88 & 10.70 & 3.62 & 2.25 & 0.65 & 0.55 & 8.86 \\
IRFS\cite{wang2023interactively}             & IF 23      & 6.64 & 30.91 & 9.88  & 2.96 & 2.18 & 0.70 & 0.59 & 9.86 \\
LRRNet\cite{li2023lrrnet}           & TPAMI 23   & 6.28 & 26.26 & 10.18 & 3.08 & 1.96 & 0.61 & 0.55 & 12.71 \\
DDFM\cite{zhao2023ddfm}             & ICCV 23    & 6.63 & 30.41 & 8.30  & 2.63 & 2.91 & 0.61 & 0.13 & 12.00 \\
SegMif\cite{liu2023multi}           & ICCV 23    & \textbf{{6.87}} & \underline{{37.42}} & 13.93 & 4.20 & 2.26 & 0.84 & \textbf{{0.69}} & 3.71 \\
CDDFuse\cite{zhao2023cddfuse}          & CVPR 23    & 6.77 & 37.06 & 14.59 & 4.30 & 3.00 & \textbf{{0.88}} & \underline{{0.68}} & \underline{{2.71}} \\
EMMA\cite{zhao2024equivariant}             & CVPR 24    & 6.77 & 36.80 & \textbf{{15.00}} & \textbf{{4.68}} & 2.80 & 0.83 & 0.65 & 3.57 \\
MMDRFuse\cite{deng2024mmdrfuse}         & ACM MM 24  & 6.35 & 26.03 & 10.23 & 3.07 & 2.40 & 0.72 & 0.58 & 10.57 \\
SFMFusion(Ours)  & -          & 6.81 & \textbf{{38.63}} & \underline{{14.66}} & \underline{{4.37}} & \underline{{3.04}} & \underline{{0.87}} & 0.67 & \textbf{{2.14}} \\
\bottomrule
\end{tabular}
\vspace{-4mm}
\end{table*}
\begin{table*}[htpb]
\caption{Quantitative results on the MRI-CT Harvard Medical dataset. The best result is in \textbf{bold} and the second-best is \underline{underlined}.}
\vspace{-2mm}
\label{tab:comparison4}
\centering
\renewcommand{\arraystretch}{1.0} 
\setlength{\tabcolsep}{8pt} 
\begin{tabular}{llcccccccccc}
\toprule
\textbf{Method} & \textbf{Venue} & \textbf{EN} ($\uparrow$) & \textbf{SD} ($\uparrow$) & \textbf{SF} ($\uparrow$) & \textbf{AG} ($\uparrow$) & \textbf{MI} ($\uparrow$) & \textbf{VIF} ($\uparrow$) & \textbf{Q\textsuperscript{AB/F}} ($\uparrow$) & \textbf{Avg.Rank} ($\downarrow$) \\
\midrule
RFN-Nest\cite{li2021rfn}         & IF 21      & 4.87 & 62.97 & 12.90 & 3.76 & 2.15 & 0.37 & 0.22 & 12.86 \\
SDNet\cite{zhang2021sdnet}            & IJCV 21    & 5.16 & 46.54 & 26.97 & 7.31 & 2.01 & 0.37 & 0.51 & 9.57 \\
SeAFusion\cite{tang2022image}        & IF 22      & 5.11 & 81.22 & 30.26 & 7.82 & 2.26 & 0.51 & 0.61 & 4.71 \\
UMF-CMGR\cite{UMF}         & IJCAI 22   & \textbf{{5.50}} & 49.76 & 30.00 & 6.52 & 2.09 & 0.36 & 0.44 & 9.57 \\
SwinFusion\cite{ma2022swinfusion}       & JAS 22     & 4.83 & 83.03 & 30.88 & 7.21 & 2.34 & \underline{{0.57}} & 0.58 & \underline{{4.57}} \\
U2Fusion\cite{xu2020u2fusion}         & TPAMI 22   & 4.50 & 44.54 & 18.92 & 5.30 & 1.96 & 0.35 & 0.38 & 14.86 \\
DeFusion\cite{liang2022fusion}         & ECCV 22    & 4.60 & 63.89 & 21.18 & 5.35 & 2.19 & 0.47 & 0.43 & 10.71 \\
ReCoNet\cite{huang2022reconet}          & ECCV 22    & 4.16 & 72.19 & 22.25 & 5.75 & 2.06 & 0.44 & 0.43 & 11.71 \\
IRFS\cite{wang2023interactively}             & IF 23      & 5.16 & 73.33 & 28.09 & 6.61 & \underline{{2.40}} & 0.45 & 0.57 & 6.00 \\
LRRNet\cite{li2023lrrnet}           & TPAMI 23   & 4.81 & 46.39 & 20.66 & 4.72 & 2.12 & 0.38 & 0.34 & 13.29 \\
DDFM\cite{zhao2023ddfm}             & ICCV 23    & 4.53 & 59.92 & 20.69 & 4.99 & \textbf{{2.58}} & 0.45 & 0.41 & 10.71 \\
SegMif\cite{liu2023multi}           & ICCV 23    & 4.29 & 81.37 & 32.44 & 7.34 & 2.30 & 0.55 & \underline{{0.62}} & 5.29 \\
CDDFuse\cite{zhao2023cddfuse}          & CVPR 23    & 4.73 & \textbf{{88.38}} & \underline{{33.82}} & \textbf{{8.07}} & 2.23 & 0.50 & 0.59 & 4.71 \\
EMMA\cite{zhao2024equivariant}             & CVPR 24    & \underline{{5.27}} & 79.60 & 27.92 & 7.26 & 2.27 & 0.49 & 0.55 & 6.14 \\
MMDRFuse\cite{deng2024mmdrfuse}         & ACM MM 24  & 4.82 & 72.08 & 24.51 & 5.82 & 2.34 & 0.50 & 0.50 & 7.86 \\
SFMFusion(Ours)  & -          & 5.07 & \underline{{86.54}} & \textbf{{35.02}} & \underline{{7.85}} & 2.31 & \textbf{{0.60}} & \textbf{{0.63}} & \textbf{{2.57}} \\
\bottomrule
\end{tabular}
\vspace{-6mm}
\end{table*}
\subsection{Evaluation Metrics}
We use eight metrics to evaluate the fusion results: Entropy (EN), Standard Deviation (SD), Spatial Frequency (SF), Average Gradient (AG), Mutual Information (MI), Visual Information Fidelity (VIF), Edge-based similarity measure (Q\textsuperscript{AB/F}) and Average Rank (Avg.Rank).
Note that, larger EN, SD, SF, AG, MI, VIF and Q\textsuperscript{AB/F} indicate better fusion performances in certain aspects.
Smaller Avg.Rank indicates a better overall fusion performance.
The details of these evaluation metrics can be found in~\cite{ma2019infrared,park2023cross}.
\subsection{Experiment Settings}
Our experiments are conducted with the PyTorch toolbox and a single NVIDIA GeForce A100 GPU.
The training samples are resized to 128$\times$128.
No data augmentation beyond resizing is applied.
The number of training epochs is set to 30.
The batch size is set to 20.
We adopt the AdamW optimizer~\cite{loshchilov2017decoupled} with the learning rate set to \( 10^{-3} \) and the weight decay set to \( 10^{-2} \).
We warm up our model for 800 iterations with a linearly growing learning rate from \( 10^{-5} \) to \( 10^{-3} \).
The exponential learning rate decay is used during the training.
For the hyper-parameters, the number of SFMGs for three branches is 3.
The number of SFMBs in each SFMG is 2.
\( \alpha_1 \),  \( \alpha_2 \), \( \alpha_3 \) and \( \alpha_4 \) are set to 0.5, 0.5, 1 and 1, respectively.
%
\subsection{Comparisons with Other Methods}
In this subsection, we compare our method with other state-of-the-art methods including RFN-Nest~\cite{li2021rfn}, SDNet~\cite{zhang2021sdnet}, SeAFusion~\cite{tang2022image}, UMF-CMGR~\cite{UMF}, SwinFusion~\cite{ma2022swinfusion}, U2Fusion~\cite{xu2020u2fusion}, DeFusion~\cite{liang2022fusion}, ReCoNet~\cite{huang2022reconet}, IRFS~\cite{wang2023interactively}, LRRNet~\cite{li2023lrrnet}, DDFM~\cite{zhao2023ddfm}, SegMif~\cite{liu2023multi}, CDDFuse~\cite{zhao2023cddfuse}, EMMA~\cite{zhao2024equivariant} and MMDRFuse~\cite{deng2024mmdrfuse}.
\subsubsection{Quantitative Comparisons}
\begin{table*}[htpb]
\caption{Quantitative results on the MRI-PET Harvard Medical dataset. The best result is in \textbf{bold} and the second-best is \underline{underlined}.}
\vspace{-2mm}
\label{tab:comparison5}
\centering
\renewcommand{\arraystretch}{1.0} 
\setlength{\tabcolsep}{8pt} 
\begin{tabular}{llcccccccccc}
\toprule
\textbf{Method} & \textbf{Venue} & \textbf{EN} ($\uparrow$) & \textbf{SD} ($\uparrow$) & \textbf{SF} ($\uparrow$) & \textbf{AG} ($\uparrow$) & \textbf{MI} ($\uparrow$) & \textbf{VIF} ($\uparrow$) & \textbf{Q\textsuperscript{AB/F}} ($\uparrow$) & \textbf{Avg.Rank} ($\downarrow$) \\
\midrule
RFN-Nest\cite{li2021rfn}         & IF 21      & 4.45 & 60.97 & 9.26  & 3.39 & 1.67 & 0.46 & 0.22 & 12.86 \\
SDNet\cite{zhang2021sdnet}            & IJCV 21    & 4.64 & 45.58 & 20.52 & 6.21 & 1.66 & 0.47 & 0.57 & 9.14 \\
SeAFusion\cite{tang2022image}        & IF 22      & \underline{{4.76}} & 73.74 & 22.06 & 6.79 & 1.81 & 0.61 & 0.62 & 4.71 \\
UMF-CMGR\cite{UMF}         & IJCAI 22   & \textbf{{4.93}} & 42.11 & 20.60 & 5.55 & 1.49 & 0.37 & 0.40 & 10.86 \\
SwinFusion\cite{ma2022swinfusion}       & JAS 22     & 4.54 & 74.91 & 22.43 & 6.78 & 1.91 & \underline{{0.70}} & \underline{{0.65}} & 3.86 \\
U2Fusion\cite{xu2020u2fusion}         & TPAMI 22   & 4.00 & 40.44 & 15.98 & 4.61 & 1.57 & 0.42 & 0.38 & 13.57 \\
DeFusion\cite{liang2022fusion}         & ECCV 22    & 4.11 & 61.00 & 22.00 & 5.97 & 1.73 & 0.56 & 0.56 & 9.14 \\
ReCoNet\cite{huang2022reconet}          & ECCV 22    & 3.16 & 59.13 & 13.61 & 4.14 & 1.44 & 0.47 & 0.30 & 13.86 \\
IRFS\cite{wang2023interactively}             & IF 23      & 4.58 & 63.25 & 19.54 & 5.61 & 1.84 & 0.58 & 0.54 & 8.00 \\
LRRNet\cite{li2023lrrnet}           & TPAMI 23   & 4.36 & 48.26 & 13.40 & 3.91 & 1.56 & 0.37 & 0.21 & 13.86 \\
DDFM\cite{zhao2023ddfm}             & ICCV 23    & 3.91 & 61.11 & 19.14 & 5.36 & \textbf{{2.12}} & 0.63 & 0.56 & 8.43 \\
SegMif\cite{liu2023multi}           & ICCV 23    & 3.93 & 72.97 & 24.76 & 7.03 & 1.79 & 0.54 & 0.61 & 6.86 \\
CDDFuse\cite{zhao2023cddfuse}          & CVPR 23    & 4.15 & \textbf{{81.49}} & \textbf{{28.04}} & \underline{{7.63}} & 1.86 & 0.66 & \underline{{0.65}} & \underline{{3.43}} \\
EMMA\cite{zhao2024equivariant}             & CVPR 24    & 4.74 & 76.04 & 22.35 & 6.74 & 1.75 & 0.59 & 0.57 & 5.57 \\
MMDRFuse\cite{deng2024mmdrfuse}         & ACM MM 24  & 4.26 & 65.42 & 18.10 & 5.27 & 1.76 & 0.59 & 0.53 & 9.43 \\
SFMFusion(Ours)  & -          & \textbf{{4.93}} & \underline{{79.09}} & \underline{{27.40}} & \textbf{{7.68}} & \underline{{2.01}} & \textbf{{0.72}} & \textbf{{0.68}} & \textbf{{1.43}} \\
\bottomrule
\end{tabular}
\vspace{-4mm}
\end{table*}
\begin{table*}[htpb]
\caption{Quantitative results on the MRI-SPECT Harvard Medical dataset. The best result is in \textbf{bold} and the second-best is \underline{underlined}.}
\vspace{-2mm}
\label{tab:comparison6}
\centering
\renewcommand{\arraystretch}{1.0} 
\setlength{\tabcolsep}{8pt} 
\begin{tabular}{llcccccccccc}
\toprule
\textbf{Method} & \textbf{Venue} & \textbf{EN} ($\uparrow$) & \textbf{SD} ($\uparrow$) & \textbf{SF} ($\uparrow$) & \textbf{AG} ($\uparrow$) & \textbf{MI} ($\uparrow$) & \textbf{VIF} ($\uparrow$) & \textbf{Q\textsuperscript{AB/F}} ($\uparrow$) & \textbf{Avg.Rank} ($\downarrow$) \\
\midrule
RFN-Nest\cite{li2021rfn}         & IF 21      & 4.02 & 60.50 & 6.41  & 2.05 & 1.78 & 0.47 & 0.19 & 12.14 \\
SDNet\cite{zhang2021sdnet}            & IJCV 21    & 4.27 & 43.53 & 16.42 & 4.57 & 1.80 & 0.57 & 0.65 & 7.14 \\
SeAFusion\cite{tang2022image}        & IF 22      & 4.28 & 60.97 & 17.25 & 4.61 & 1.87 & 0.60 & 0.65 & 4.29 \\
UMF-CMGR\cite{UMF}         & IJCAI 22   & \underline{{4.52}} & 33.90 & 14.39 & 3.65 & 1.66 & 0.40 & 0.45 & 11.00 \\
SwinFusion\cite{ma2022swinfusion}       & JAS 22     & 4.11 & 60.11 & 16.31 & 4.25 & 1.92 & 0.63 & 0.63 & 5.43 \\
U2Fusion\cite{xu2020u2fusion}         & TPAMI 22   & 3.62 & 36.39 & 12.57 & 3.23 & 1.70 & 0.47 & 0.45 & 13.29 \\
DeFusion\cite{liang2022fusion}         & ECCV 22    & 3.73 & 51.33 & 14.55 & 3.60 & 1.82 & 0.60 & 0.55 & 9.43 \\
ReCoNet\cite{huang2022reconet}          & ECCV 22    & 2.96 & 43.28 & 9.59  & 2.61 & 1.39 & 0.36 & 0.27 & 14.86 \\
IRFS\cite{wang2023interactively}             & IF 23      & 4.21 & 54.24 & 13.74 & 3.54 & 1.92 & 0.58 & 0.53 & 8.29 \\
LRRNet\cite{li2023lrrnet}           & TPAMI 23   & 4.00 & 42.24 & 10.88 & 2.90 & 1.64 & 0.34 & 0.20 & 13.86 \\
DDFM\cite{zhao2023ddfm}             & ICCV 23    & 3.77 & 59.18 & 14.71 & 3.77 & \textbf{{2.11}} & 0.61 & 0.63 & 6.57 \\
SegMif\cite{liu2023multi}           & ICCV 23    & 3.63 & \underline{{67.74}} & 16.84 & 4.10 & 1.77 & 0.58 & 0.59 & 7.71 \\
CDDFuse\cite{zhao2023cddfuse}          & CVPR 23    & 3.82 & \textbf{{71.62}} & \textbf{{20.66}} & \underline{{5.15}} & 1.89 & \underline{{0.65}} & \underline{{0.68}} & \underline{{3.43}} \\
EMMA\cite{zhao2024equivariant}             & CVPR 24    & 4.44 & 65.31 & 17.31 & 4.54 & 1.83 & 0.55 & 0.58 & 6.00 \\
MMDRFuse\cite{deng2024mmdrfuse}         & ACM MM 24  & 3.87 & 55.61 & 13.53 & 3.44 & 1.84 & 0.58 & 0.52 & 9.71 \\
SFMFusion(Ours)  & -          & \textbf{{4.68}} & 65.46 & \underline{{20.48}} & \textbf{{5.26}} & \underline{{1.98}} & \textbf{{0.72}} & \textbf{{0.74}} & \textbf{{1.57}} \\
\bottomrule
\end{tabular}
\vspace{-6mm}
\end{table*}
For IVIF, we present the comparative results in Tab.~\ref{tab:comparison1}, \ref{tab:comparison2} and \ref{tab:comparison3}.
Our method achieves the best Avg.Rank on three datasets, demonstrating that our method outperforms other methods.
Noted that, CDDFuse~\cite{zhao2023cddfuse} and EMMA~\cite{zhao2024equivariant} show comparable results, which combine the advantages of CNNs and Transformers.
Different from them, we utilize spatial-frequency enhanced Mamba to extract comprehensive features.
Thus, our method has a better Avg.Rank.

For MIF, we present the comparative results in Tab.~\ref{tab:comparison4}, \ref{tab:comparison5} and \ref{tab:comparison6}.
Our method achieves the best Avg.Rank on three datasets.
Noted that, both VIF and Q\textsuperscript{AB/F} reach the optimal values.
This indicates that our method preserves more information from source images, which is attributed to our IR branches and the dynamic feature fusion strategy.
On the MRI-CT Harvard Medical dataset, our method achieves relatively lower EN and MI values.
The main reason is that the MRI-CT dataset has the largest domain gap among the six datasets.
These results suggest that the generalization ability of our method still requires further improvement.

In the field of MMIF, due to the lack of ground truth, the evaluation is multifaceted.
However, in this situation, our method still achieves nearly optimal and suboptimal results across eight metrics on six datasets.
The main reason is that our method enhances Mamba in the spatial-frequency domains and dynamically fuse information from the IR branches.
\subsubsection{Qualitative Comparisons}
\begin{figure*}[!t]
  \centering
  \includegraphics[width=0.96\textwidth]{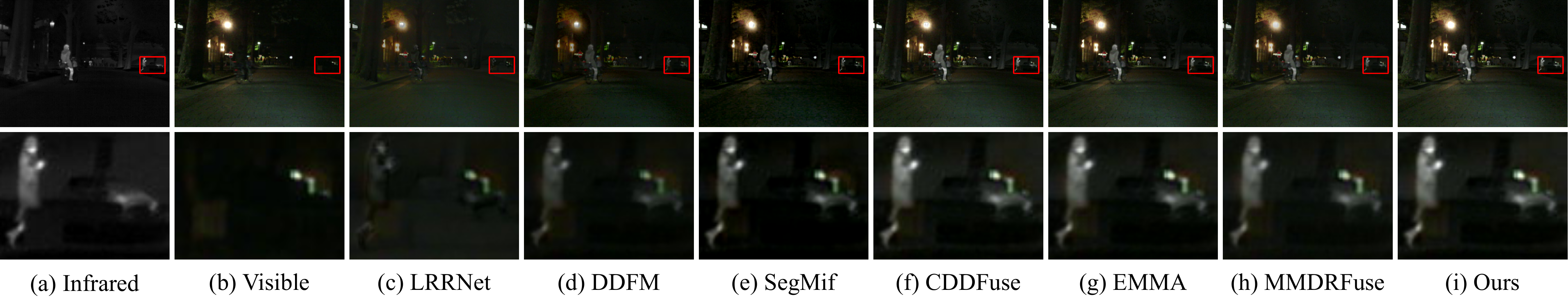}
  \vspace{-4mm}
  \caption{Comparison of fusion results on the MSRS dataset. The second row shows the enlarged regions in the first row.}
  \label{fig:compare1}
  \vspace{-2mm}
\end{figure*}

\begin{figure*}[htpb]
  \centering
  \includegraphics[width=0.96\textwidth]{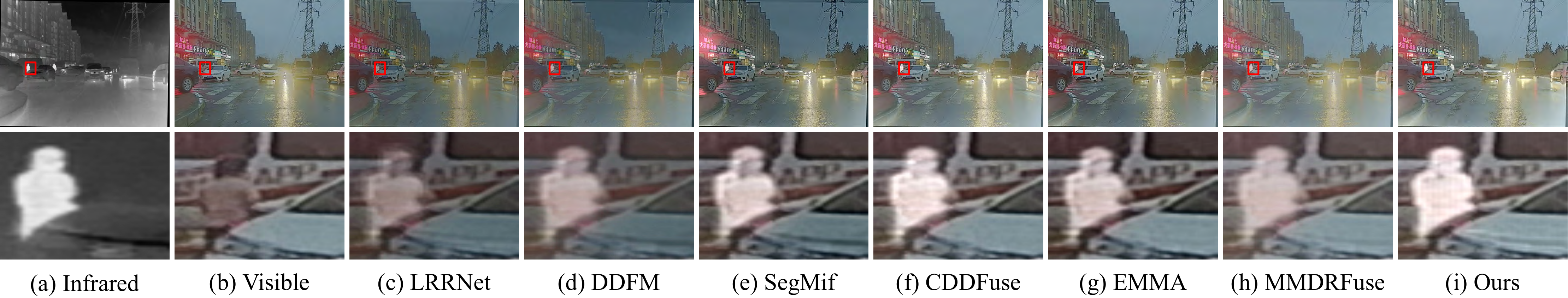}
  \vspace{-4mm}
  \caption{Comparison of fusion results on the M3FD dataset. The second row shows the enlarged regions in the first row.}
  \label{fig:compare2}
  \vspace{-2mm}
\end{figure*}

\begin{figure*}[htpb]
  \centering
  \includegraphics[width=0.96\textwidth]{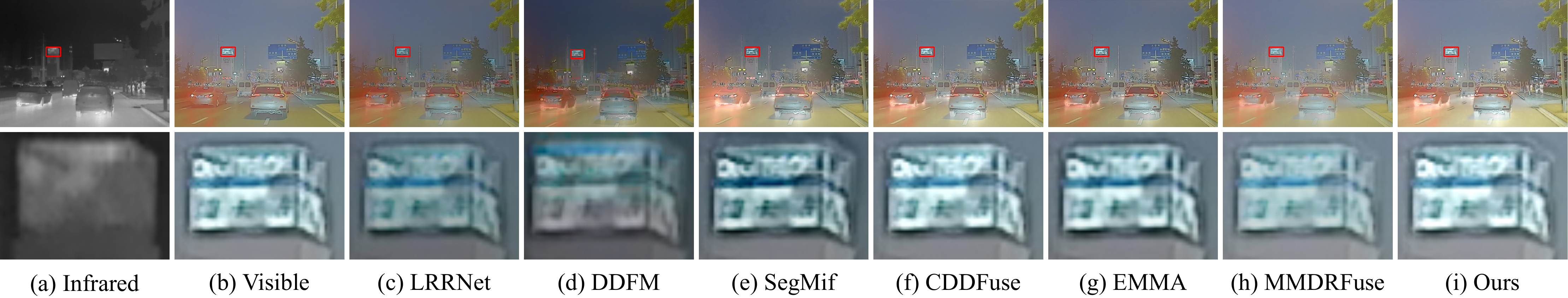}
  \vspace{-4mm}
  \caption{Comparison of fusion results on the FMB dataset. The second row shows the enlarged regions in the first row.}
  \label{fig:compare3}
  \vspace{-2mm}
\end{figure*}

For IVIF, we show the qualitative comparison in Fig.~\ref{fig:compare1}, \ref{fig:compare2} and \ref{fig:compare3}.
Our method demonstrates a superior integration of thermal radiation information from infrared images and detailed textures from visible images.
As shown in Fig.~\ref{fig:compare1} and \ref{fig:compare2}, objects in dark regions are prominently highlighted, allowing foreground targets to be easily distinguished from the background.
These results indicate that our SFMFusion could potentially benefit subsequent high-level vision tasks, such as object detection and tracking.
As shown in Fig.~\ref{fig:compare3}, background details, which are typically difficult to discern under low illumination conditions, are rendered with clear edges and rich contour information.

\begin{figure*}[htpb]
  \centering
  \includegraphics[width=0.96\textwidth]{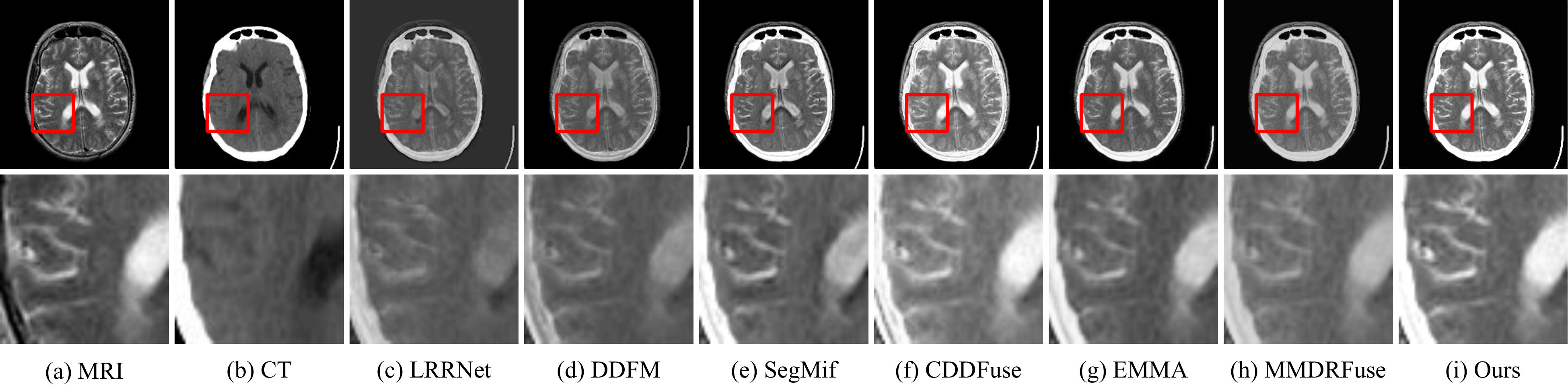}
  \vspace{-4mm}
  \caption{Comparison of fusion results on the MRI-CT Harvard Medical dataset. The second row shows the enlarged regions in the first row.}
  \label{fig:compare4}
  \vspace{-2mm}
\end{figure*}

\begin{figure*}[htpb]
  \centering
  \includegraphics[width=0.96\textwidth]{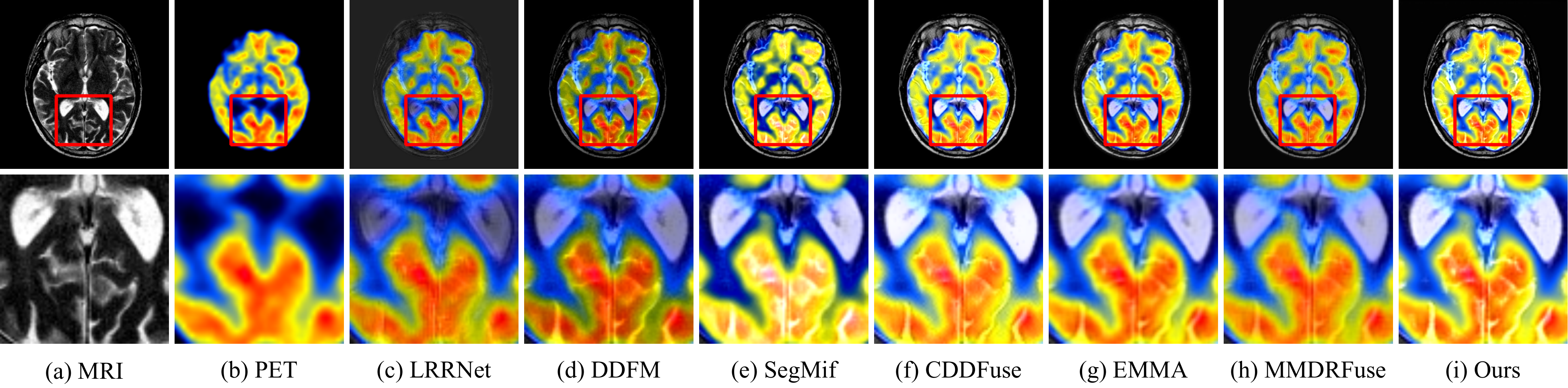}
  \vspace{-4mm}
  \caption{Comparison of fusion results on the MRI-PET Harvard Medical dataset. The second row shows the enlarged regions in the first row.}
  \label{fig:compare5}
  \vspace{-2mm}
\end{figure*}

\begin{figure*}[htpb]
  \centering
  \includegraphics[width=0.96\textwidth]{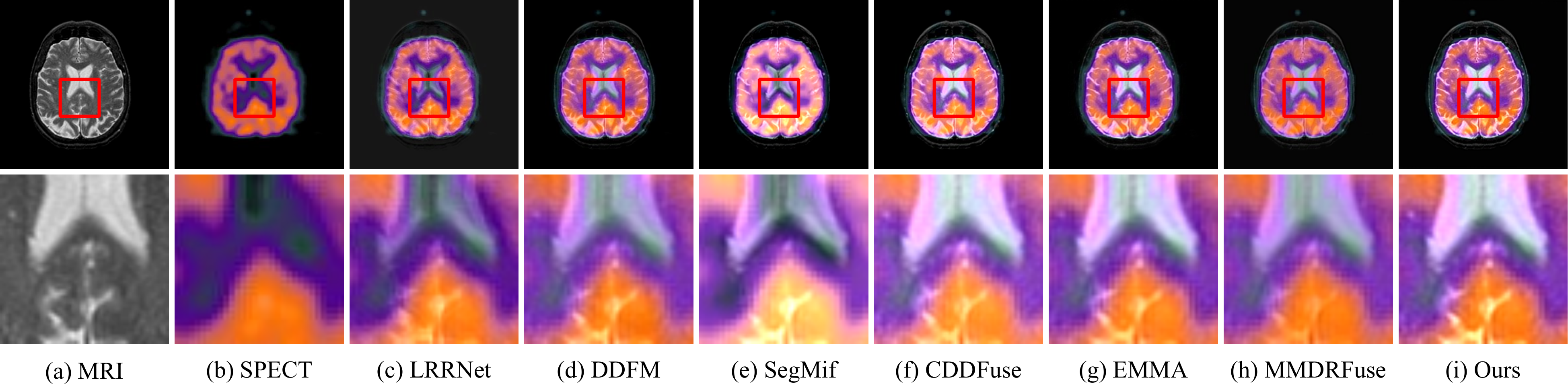}
  \vspace{-4mm}
  \caption{Comparison of fusion results on the MRI-SPECT Harvard Medical dataset. The second row shows the enlarged regions in the first row.}
  \label{fig:compare6}
  \vspace{-2mm}
\end{figure*}

For MIF, we show the qualitative comparison in Fig.~\ref{fig:compare4}, \ref{fig:compare5} and \ref{fig:compare6}.
Our method fully integrates the detailed information of MRI images and the functional information of the other images.
As shown in Fig.~\ref{fig:compare4}, our method clearly demonstrates the detailed information of the MRI images.
As shown in Fig.~\ref{fig:compare5} and \ref{fig:compare6}, our method not only preserves the color information but also achieves higher contrast.
This makes it much easier to observe the images for better medical diagnosis.

In summary, the qualitative comparisons for both IVIF and MIF clearly demonstrate the effectiveness of our method in integrating multi-modal information.
\subsection{Ablation Study}
In this subsection, ablation studies are conducted on the MSRS~\cite{tang2022piafusion} dataset.
EN, SD, SF, AG and Avg.Rank are used to quantitatively validate the fusion performance.
\subsubsection{Effectiveness of Key Components}
We conduct experiments to verify the effectiveness of the proposed key components.
The results are shown in Tab.~\ref{tab:ablation}.
\begin{table}[htpb]
\caption{Ablation results of key components on MSRS.}
\vspace{-3mm}
\label{tab:ablation}
\centering
\renewcommand{\arraystretch}{1.0} 
\setlength{\tabcolsep}{2pt} 
\begin{tabular}{llccccc}
\toprule
 & \textbf{Configurations} & \textbf{EN} ($\uparrow$) & \textbf{SD} ($\uparrow$) & \textbf{SF} ($\uparrow$) & \textbf{AG} ($\uparrow$) & \textbf{Avg. Rank} ($\downarrow$) \\
\midrule
I   & w/o IR & 6.65 & 42.07 & 11.26 & 3.68 & 4.25 \\
II  & w/o CEB                  & 6.67 & 42.53 & 11.36  & 3.71 & 2.25 \\
III & w/o FEB                  & 6.62 & 41.08 & 11.03 & 3.56 & 7.25 \\
IV  & 2-scale MMB    & 6.67 & 41.99 & 11.16 & 3.63 & 4.50 \\
V & 1-scale MMB    & 6.61 & 41.80 & 11.30 & 3.64 & 5.50 \\
VI  & DFMB $\rightarrow$ Add     & 6.65 & 42.73 & 11.30 & 3.60 & 4.00 \\
VII   & DFMB $\rightarrow$ Concat   & 6.66 & 41.81 & 11.01 & 3.53 & 6.50 \\
 & Ours              & \textbf{6.72} & \textbf{45.17} & \textbf{11.94} & \textbf{3.90} & \textbf{1.00} \\
\bottomrule
\end{tabular}
\vspace{-2mm}
\end{table}
\begin{figure}[t]
  \centering
  \includegraphics[width=0.9\linewidth]{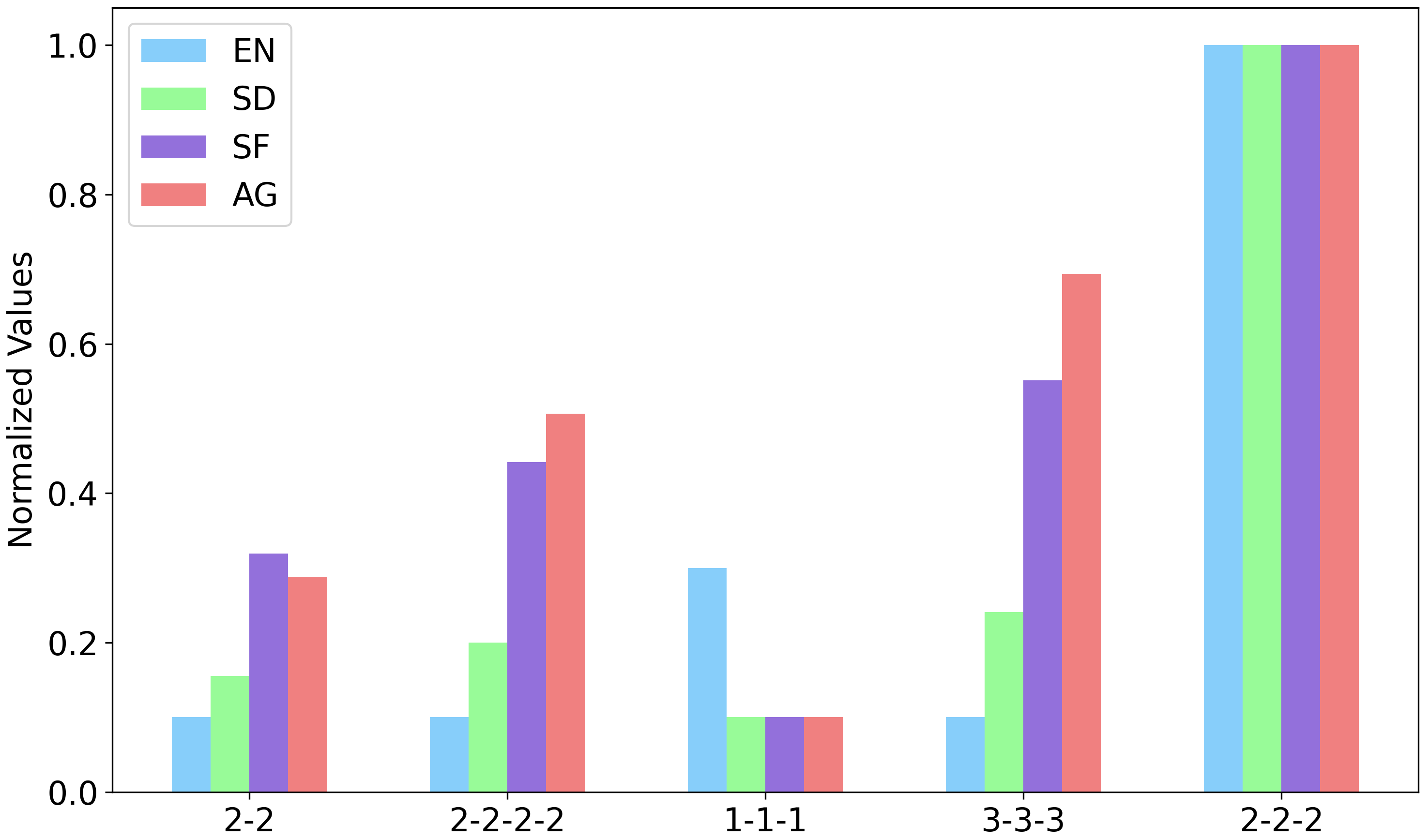}
  \vspace{-2mm}
  \caption{Effects of stacked SFMG and SFMB.}
  \label{fig:Stack}
  \vspace{-4mm}
\end{figure}

\textbf{Effects of Image Reconstruction.}
In Exp.I, we remove the IR task by simultaneously removing the last two convolutional layers and and the overall residual connections of the two IR branches.
Without the IR task, all the evaluation metrics fail to match the performance of our SFMFusion.
The experimental results show that IR is one key to the excellent performance of our SFMFusion.

\textbf{Effects of SFMB.}
In Exp.II and III, we remove CEB and FEB separately.
It can be seen that removing either block results in a decrease in all metrics.
Notably, removing FEB leads to a significant decline in SF and AG.
These metrics are closely related to image texture details.
This indicates the critical role of FEB in handling high-frequency information.
In Exp.IV and V, we replace MMB with a 2-scale MMB and an 1-scale MMB.
The experimental results show that the 2-scale MMB performs slightly better than the 1-scale MMB.
However, the 3-scale MMB (Ours) achieves significant improvements across all metrics.
Therefore, we choose the 3-scale MMB as the final model.

\textbf{Effects of DFMB.}
In Exp.VI and VII, we replace DFMB by directly adding or concatenating the features of the three branches.
The experimental results show that these simple fusion strategies fail to fully utilize features from three different branches.
All metrics are decreased due to these improper feature fusion strategies.
\subsubsection{Effects of Stacked SFMG and SFMB}
We conduct experiments to verify the effects of stacked SFMG and SFMB.
The results are shown in Fig.~\ref{fig:Stack}.
Here, 2-2 indicates that there are two SFMGs, each containing two SFMBs.

\textbf{Effects of Stacked SFMG.}
The performance improves as the number of the stacked SFMGs increases from 2 to 3. This shows that adding more groups helps the model capture more comprehensive features. However, when the number of groups exceeds 3, the performance decreases, indicating that further stacking groups provides limited benefits.

\textbf{Effects of Stacked SFMB.}
Similarly, the performance improves as the number of the stacked SFMBs increases from 1 to 2.
However, stacking more than 2 SFMBs leads to a decrease in performance.
Thus, we set 2-2-2 by default.
\subsubsection{Effects of Input Resolution and Feature Dimension}
\begin{figure}[tbp]
  \centering
  \includegraphics[width=\linewidth]{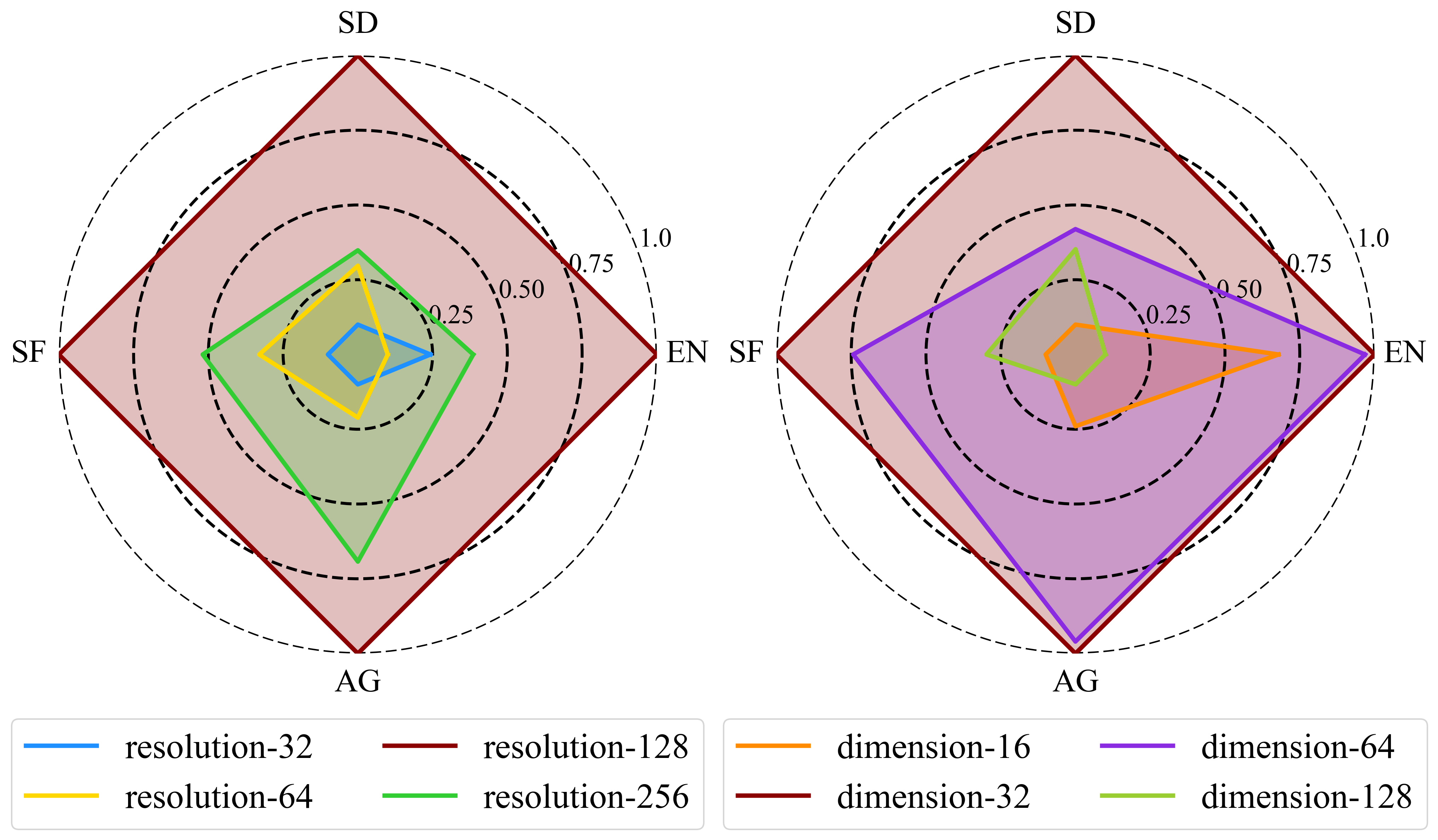}
  \vspace{-6mm}
  \caption{Effects of input resolution and feature dimension.}
  \label{fig:Resolution_Channel}
  \vspace{-6mm}
\end{figure}

We conduct experiments to verify the effects of input resolution and feature dimension.
The corresponding results can be found in Fig.~\ref{fig:Resolution_Channel}.
Here, resolution-32 indicates that the input resolution is 32×32, and dimension-16 indicates that the feature dimension is 16 in the SFMB.

\textbf{Effects of Input Resolution.}
We change the input resolutions from 32$\times$32 to 256$\times$256.
When the input resolution is 32$\times$32, the metrics show worst values.
This is because the excessively low resolution results in the loss of image details.
When the input resolution is 128$\times$128, all the metrics have reached their optimal values.
When the input resolution is 256$\times$256, all the metrics exhibit different levels of reduction.
Thus, we set 128$\times$128 as the default input resolution.

\textbf{Effects of Feature Dimension.}
We change the feature dimensions from 16 to 128.
We find that the performance is worst when the dimension is 16, and the performance is best when the dimension is 32.
With the dimensions increase, the performance begins to decrease.
The reason may be that the model tends to over-fitting when the dimension is too high.
Thus, we set 32 as the default feature dimension.
\begin{table}[t]
\caption{Results with different input preprocessing methods on MSRS.}
\vspace{-3mm}
\label{tab:random crop}
\centering
\renewcommand{\arraystretch}{1.0} 
\setlength{\tabcolsep}{2pt} 
\begin{tabular}{llccccc}
\toprule
 & \textbf{Configurations} & \textbf{EN} ($\uparrow$) & \textbf{SD} ($\uparrow$) & \textbf{SF} ($\uparrow$) & \textbf{AG} ($\uparrow$) & \textbf{Avg. Rank} ($\downarrow$) \\
\midrule
I   & Random Crop 32$\times$32 & 6.66 & 42.25 & 11.19 & 3.64 & 4.75 \\
II  & Random Crop 64$\times$64 & 6.66 & 42.62 & 11.46  & 3.77 & 3.50 \\
III & Random Crop 128$\times$128 & 6.67 & 42.55 & 11.57 & 3.83 & 2.75 \\
IV  & Random Crop 256$\times$256 & 6.67 & 42.48 & 11.65 & 3.86 & 2.50 \\
 & Resize 128$\times$128(Ours) & \textbf{6.72} & \textbf{45.17} & \textbf{11.94} & \textbf{3.90} & \textbf{1.00} \\
\bottomrule
\end{tabular}
\vspace{-5mm}
\end{table}

\textbf{Effects of Input Preprocessing.}
We compare our resizing strategy with random cropping.
The results are shown in Tab.~\ref{tab:random crop}.
It shows that larger random cropping sizes lead to better performance.
However, the performance remains consistently inferior to our resizing strategy.
The main reason is that the random cropping discards global information, which hinders Mamba’s ability to model global dependencies.
\begin{table}[t]
\caption{Results with different loss function weights on MSRS.}
\vspace{-3mm}
\label{tab:loss weight}
\centering
\renewcommand{\arraystretch}{1.0} 
\setlength{\tabcolsep}{2pt} 
\begin{tabular}{llcccc}
\toprule
 & \textbf{Configurations} & \textbf{EN} ($\uparrow$) & \textbf{SD} ($\uparrow$) & \textbf{SF} ($\uparrow$) & \textbf{AG} ($\uparrow$) \\
\midrule
I   & \( \alpha_1 \)=1.0,  \( \alpha_2 \)=1.0, \( \alpha_3 \)=1.0, \( \alpha_4 \)=1.0 & 6.64 & 42.39 & 11.39 & 3.70 \\
II  & \( \alpha_1 \)=2.0,  \( \alpha_2 \)=2.0, \( \alpha_3 \)=1.0, \( \alpha_4 \)=1.0 & 6.68 & 42.98 & 11.37  & 3.70 \\
III & \( \alpha_1 \)=0.5,  \( \alpha_2 \)=0.5, \( \alpha_3 \)=10.0, \( \alpha_4 \)=10.0 & 6.63 & 42.01 & 11.04 & 3.51 \\
IV  & \( \alpha_1 \)=0.5,  \( \alpha_2 \)=0.5, \( \alpha_3 \)=0.1, \( \alpha_4 \)=0.1 & 6.67 & 43.06 & 11.21 & 3.57 \\
 & \( \alpha_1 \)=0.5,  \( \alpha_2 \)=0.5, \( \alpha_3 \)=1.0, \( \alpha_4 \)=1.0(Ours) & \textbf{6.72} & \textbf{45.17} & \textbf{11.94} & \textbf{3.90} \\
\bottomrule
\end{tabular}
\vspace{-4mm}
\end{table}
\begin{table}[t]
\caption{Results with different training strategies on MSRS.}
\vspace{-2mm}
\label{tab:training}
\centering
\renewcommand{\arraystretch}{1.0}
\setlength{\tabcolsep}{2pt}
\begin{tabular}{llcccc}
\toprule
 & \textbf{Configurations} & \textbf{EN} ($\uparrow$) & \textbf{SD} ($\uparrow$) & \textbf{SF} ($\uparrow$) & \textbf{AG} ($\uparrow$) \\
\midrule
I   & Two-Stage Training & 6.49 & 43.42 & 11.19 & 3.39 \\
II  & One-Stage Training with Task Head & 6.63 & 42.62 & 11.21  & 3.57 \\
 & Ours & \textbf{6.72} & \textbf{45.17} & \textbf{11.94} & \textbf{3.90} \\
\bottomrule
\end{tabular}
\vspace{-4mm}
\end{table}
\begin{table}[t]
\caption{Comparison of computational efficiency on M3FD.}
\vspace{-3mm}
\label{tab:complexity}
\centering
\renewcommand{\arraystretch}{1.0} 
\setlength{\tabcolsep}{2pt} 
\begin{tabular}{llccc}
\toprule
 & \textbf{Method} & \textbf{Time(ms)} & \textbf{FLOPS(G)} & \textbf{Params(M)} \\
\midrule
I   & SeAFusion\cite{tang2022image} & 24 & 131 & 0.167 \\
II  & SwinFusion\cite{ma2022swinfusion} & 63 & 185 & 0.955 \\
III  & SegMif\cite{liu2023multi} & 682 & 500 & 0.621 \\
IV  & CDDFuse\cite{zhao2023cddfuse} & 464 & 1402 & 1.186 \\
V  & EMMA\cite{zhao2024equivariant} & 26 & 106 & 1.516 \\
VI  & DDFM\cite{zhao2023ddfm} & 280.8k & 1336k & 552.7 \\
 & Ours & 2053 & 439 & 0.766 \\
\bottomrule
\end{tabular}
\vspace{-2mm}
\end{table}
\subsubsection{Effects of Loss Function Weights and Training Strategies}
In this work, we adopt multiple loss functions and one-stage training strategies.
To evaluate the effects, we conduct more experiments on these aspects.

\textbf{Effects of Loss Function Weights.}
We conduct a detailed search on the loss function weights.
The results are reported in Tab.~\ref{tab:loss weight}.
As observed, different weight settings have a significant impact on performance.
This confirms that appropriate loss function weights are crucial for fully exploiting the capability of our model.

\textbf{Effects of Training Strategies.}
As shown in Fig.~\ref{fig:introduction}(b), there are different training strategies for image fusion models.
Tab.~\ref{tab:training} shows the effect of different training strategies.
With the same backbone, we compare our method with two-stage training and one-stage training with task head.
The results show that our one-stage training strategy achieves better overall performance, directly demonstrating its effectiveness.
\subsection{Computational Efficiency Analysis}
In this section, we evaluate computational efficiency using three widely adopted metrics: average runtime, FLOPs and parameter count.
Following the protocol in~\cite{liu2024infrared}, we randomly select 10 image pairs from the M3FD dataset, each with a resolution of 1024$\times$768, and conduct the experiments on an Nvidia GeForce 4090 GPU.
We compare our method with several representative approaches that perform well on the M3FD dataset.
As shown in Tab.~\ref{tab:complexity}, our method maintains a moderate level of FLOPs and parameters.
Although the runtime is relatively longer, our method achieves superior performance as reported in Tab.~\ref{tab:comparison2}.
\begin{figure}[!h]
  \centering
  \includegraphics[width=0.8\linewidth,height=3.6cm]{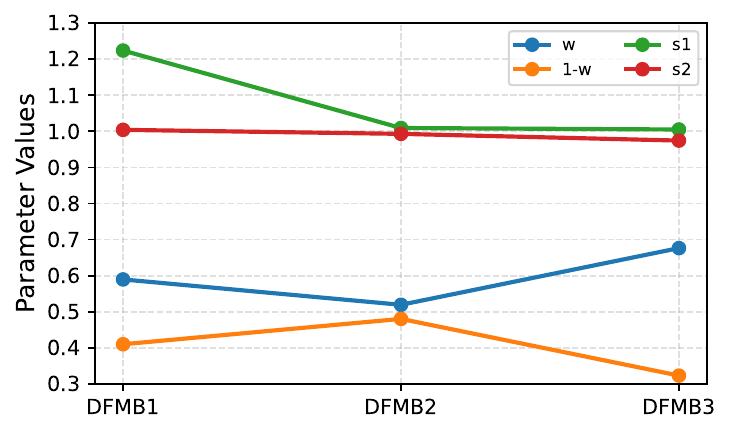}
  \vspace{-2mm}
  \caption{Visualization of DFMB weights. DFMB1 denotes the first DFMB in the framework.}
  \label{fig:weight}
  \vspace{-4mm}
\end{figure}
\begin{figure}[t]
  \centering
  \includegraphics[width=0.8\linewidth,height=4cm]{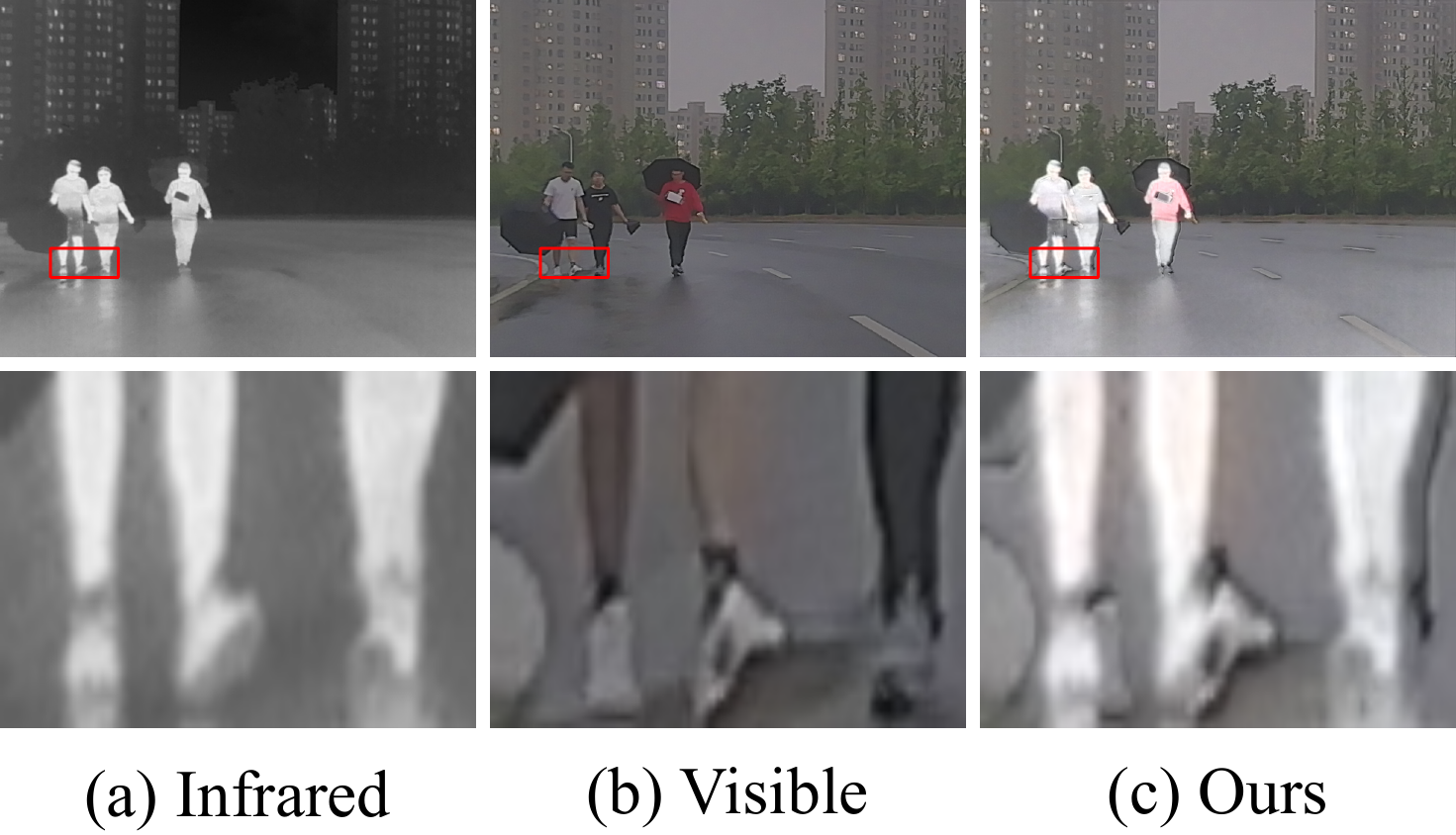}
  \vspace{-2mm}
  \caption{Visualization of the failure case. The second row shows the enlarged regions in the first row.}
  \label{fig:Failure}
  \vspace{-4mm}
\end{figure}
\subsection{Visualization and Analysis of DFMB's Parameters}
Fig. 15 visualizes the learned parameters (i.e., $W$, $1-W$, $s_1$ and $s_2$) in our proposed DFMB.
One can observe that the scaling factors $s_1$ and $s_2$ generally remain close to 1 with minor fluctuations.
The weights $W$ vary more significantly to dynamically balance different branches.
These results clearly reveal the adaptive feature fusion mechanism of DFMB and confirm its effectiveness.
\subsection{Limitation and Discussion}
Although achieving robust performance across diverse scenarios, our method also has some limitations.
As shown in Fig.~\ref{fig:Failure}, the fused image produced by our method exhibits noticeable ghosting artifacts around the legs of the person.
The main reason is that paired images are not fully aligned.
To address this problem, potential remedies can be considered, such as jointly optimizing registration and fusion or introducing supervision from high-level vision tasks.
\section{Conclusion and future work}
In this work, we propose a new feature learning framework named SFMFusion for MMIF.
Technically, we first develop a three-branch structure with the help of IR to preserve complete contents.
Then, we introduce the Spatial-Frequency Enhanced Mamba Block (SFMB) to enhance Mamba in the spatial-frequency domains.
Finally, we utilize the Dynamic Fusion Mamba Block (DFMB) to integrate the features of different branches.
On six MMIF benchmarks, our SFMFusion performs favorably against state-of-the-art methods quantitatively and qualitatively.
In future work, we plan to extend our method to other image fusion tasks, such as multi-focus image fusion and multi-exposure image fusion.
\bibliographystyle{IEEEtran}
\bibliography{IEEEabrv,main}
\end{document}